\documentclass[10pt,twocolumn,letterpaper]{article}

\usepackage[pagenumbers]{wacv} 

\usepackage{amsfonts}
\usepackage{algorithm}
\usepackage{algpseudocode}
\usepackage{hhline}
\usepackage{array}
\usepackage[normalem]{ulem}
\usepackage{multirow}

\newcommand{\mat}[1]{\mathbf{#1}}

\def\Rotation{{\mat{R}}}
\def\Translation{{\vec{t}}}
\def\Intrinsics{\mat{K}}
\def\ProjMatrix{{\mat{P}}}

\def\yawAngle{{\theta}}
\newcommand{\yaw}[1]{\yawAngle_{#1}}
\newcommand{\bboxGeneric}[1]{\vec{b}^{#1}}
\newcommand{\bboxImg}[1]{\vec{b}_{#1}}
\newcommand{\bboxLidar}[1]{\vec{B}_{#1}}

\renewcommand{\vec}[1]{\mathbf{#1}}

\definecolor{wacvblue}{rgb}{0.21,0.49,0.74}
\usepackage[pagebackref,breaklinks,colorlinks,allcolors=wacvblue]{hyperref}

\def\wacvPaperID{3504} 
\def\confName{WACV}
\def\confYear{2027}

\title{MiCaDe: Online LiDAR–Camera Miscalibration Change Detection for Autonomous Driving}

\author{
Carlo Sgaravatti\textsuperscript{1,*} \qquad
Riccardo Pieroni\textsuperscript{2} \qquad
Matteo Corno\textsuperscript{1} \qquad
Sergio M. Savaresi\textsuperscript{1} \\ 
Luca Magri\textsuperscript{1} \qquad 
Giacomo Boracchi\textsuperscript{1} \\
\textsuperscript{1}Politecnico di Milano \qquad
\textsuperscript{2}Niulinx\\
}

\begin{document}
\maketitle
{\def\thefootnote{}\footnotetext{\scriptsize\textsuperscript{*}Corresponding author: \texttt{carlo.sgaravatti@polimi.it}}
\begin{abstract}
Reliable multi-sensor perception in autonomous vehicles requires accurate calibration between cameras and LiDAR. However, calibration drift is often unavoidable in long-term deployments: vibrations, thermal effects, and minor mechanical shocks may alter the relative pose between sensors, leading to systematic errors in downstream fusion tasks.
We propose MiCaDe (Miscalibration Change Detection), an online solution detecting when the extrinsic LiDAR-camera calibration parameters are no longer accurate by means of sequential change detection. MiCaDe requires no calibration targets, or modification of the perception stack, providing low false-alarm rates with negligible computational overhead. Our key insight is that 3D and 2D bounding boxes produced by standard object detectors should remain geometrically consistent under correct calibration. By matching cross-modal detections, we derive geometric consistency metrics that serve as sensitive indicators of extrinsic drift. These metrics are aggregated into fixed-size batches and scored against a reference distribution, producing a likelihood stream that is monitored by a Change Point Model (CPM). Crucially, the CPM accumulates statistical evidence over time, enabling detection of subtle drifts that would otherwise go unnoticed. Experiments on multiple long-sequence datasets and on a real-world dataset with physically induced miscalibration confirm reliable detection under diverse conditions. Code available at: \url{https://github.com/CarloSgaravatti/MiCaDe}.
\end{abstract}
    
\section{Introduction}
\label{sec:intro}

\begin{figure}[t]
\centering
\includegraphics[width=\linewidth]{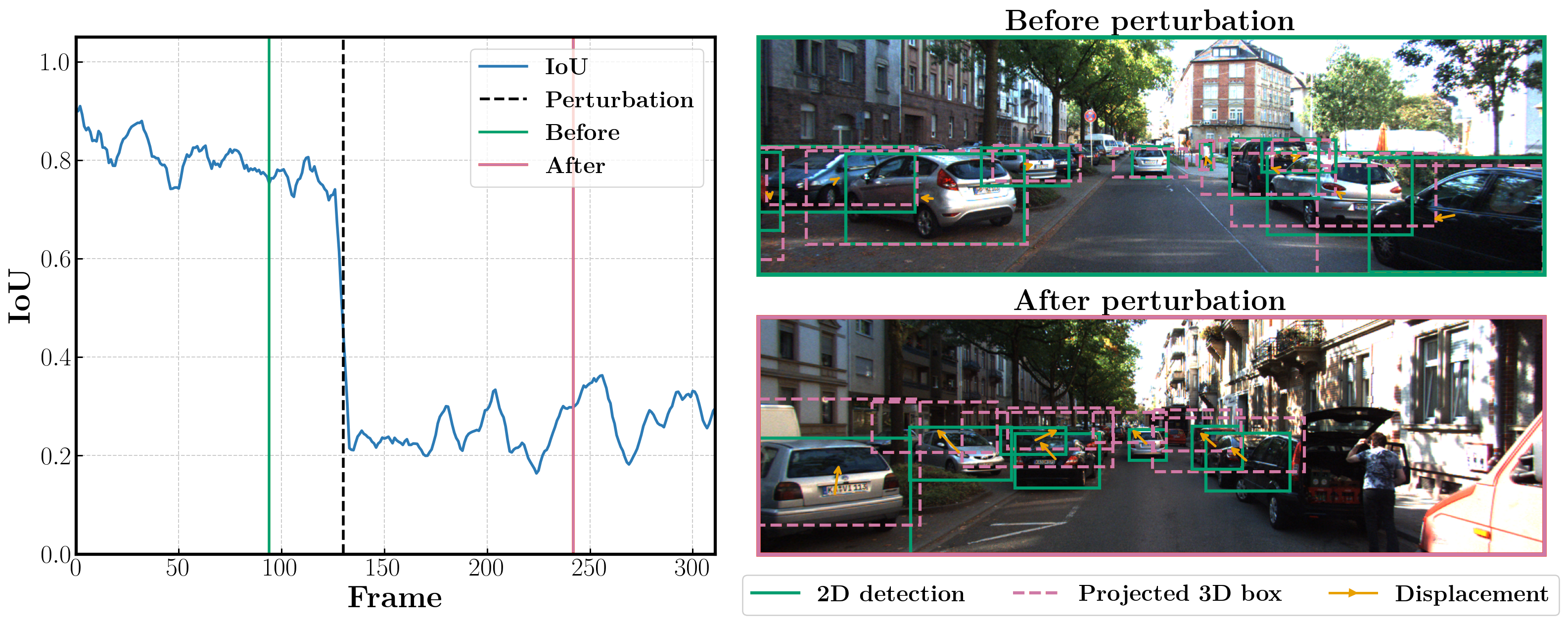}
\caption{\textbf{Calibration drift detection as a change-detection problem.}
After a perturbation of the camera extrinsics (black dashed line), the geometric consistency between projected 3D detections (red) and image detections (green) exhibits a degradation in alignment between them, as shown in their IoU (one of the monitored metrics). MiCaDe detects such changes by jointly monitoring multiple geometric consistency metrics over time.}
\label{fig:teaser}
\end{figure}

Reliable perception in autonomous driving relies on accurate fusion between complementary sensors such as cameras and LiDARs, which in turn requires a precise \emph{extrinsic calibration}. Unfortunately, minor shocks and gradual effects, such as vibrations or temperature changes, can cause calibration drifts that accumulate over time, degrading perception accuracy and compromising navigation safety.

Detecting when calibration becomes unreliable is therefore critical for long-term autonomous operation. Existing approaches typically formulate miscalibration detection as a frame-level classification problem. They first extract cross-modal cues, such as image edges, depth discontinuities, or learned multimodal features, and then train a classifier to decide whether the sensor pair is calibrated or not \cite{cross-modal-matching,tahiraj2025cal}. While effective, these approaches present several limitations. First, they usually require dedicated training procedures and synthetic calibration perturbations. Second, they operate independently on each frame, making it difficult to distinguish transient detection noise from persistent calibration drift. Third, they often require specific neural networks, introducing additional computational overhead.

In this work, we introduce \mbox{MiCaDe} (Miscalibration Change Detection), an online monitoring framework for LiDAR-Camera extrinsic calibration. We formulate calibration monitoring as a sequential change-detection problem. Rather than learning whether a sensor pair is calibrated, we explicitly model the distribution of indicators derived from signals that are already available in modern off-the-shelf perception stacks, and monitor these over time. 
Specifically, we exploit the correspondence between 3D LiDAR detections and 2D image detections of the same object. Under correct calibration, the geometric relationship between these detections remains statistically stable over time. When the calibration drifts, we expect a systematic change in the consistency metrics, as illustrated in Figure~\ref{fig:teaser}. We therefore continuously monitor the stream of geometric consistency metrics and detect statistically significant deviations from their nominal behaviour. This approach offers several advantages: it requires no task-specific training, accumulates evidence over time to detect subtle relative displacements of the sensors that may be imperceptible in individual frames, and provides control over the nominal false-alarm rate through the Average Run Length (ARL$_0$) parameter, though this control is only approximate when the monitored signal is not perfectly stationary, as we discuss in Section~\ref{sec:experiments}. To the best of our knowledge, this is the first approach that casts miscalibration detection as a sequential change-detection problem, rather than as per-frame supervised classification.

MiCaDe matches 3D and 2D detections, extracts geometric consistency indicators from the resulting correspondences, and aggregates them into fixed-size batches. A reference distribution fit on calibrated sequences is then used to score incoming observations, while a sequential change-detection test monitors their evolution and raises an alarm when a statistically significant distributional shift is detected. Since MiCaDe relies exclusively on detections already produced by the perception stack, it introduces negligible computational overhead and can be readily integrated into existing autonomous driving systems~\cite{sgaravatti2024multimodal,sgaravatti2026lcf3d}.

Our main contributions are:

\begin{itemize}
    \item \textbf{Sequential monitoring of calibration consistency:} we formulate LiDAR–camera miscalibration detection as a statistical change-detection problem, monitoring a data stream of cross-modal geometric consistency metrics and providing low false-alarm rates.
    \item \textbf{Lightweight calibration monitoring from standard detections:} we show that geometric consistency metrics derived from 2D and 3D object detections already available in modern perception stacks provide effective indicators of calibration drift, without requiring additional sensors, supervision, or retraining.
    \item \textbf{Real-world evaluation with induced drift:} we provide the first public on-road dataset with physically induced miscalibration at a known time, enabling quantitative evaluation under real conditions.
\end{itemize}
\section{Related Work}
\label{sec:related-work}

Calibration monitoring lies at the intersection of LiDAR-camera calibration and change detection. Existing approaches mainly differ in the signal used to assess cross-modal consistency and in the decision rule adopted to detect calibration failures.

\subsection{Calibration Signals}
Levinson and Thrun~\cite{Levinson2013AutomaticOC} compare image edges with LiDAR depth discontinuities to identify calibration degradation. More recent methods, instead, employ learned cross-modal representations. Chen~\emph{et al.}~\cite{cross-modal-matching} learn a consistency score through contrastive learning using a Siamese architecture applied to RGB images and depth maps obtained by projecting the LiDAR points into the image, while Tahiraj~\emph{et al.}~\cite{tahiraj2025cal} use modality-specific encoders on RGB images and depth maps to extract calibration-sensitive features.

In contrast, MiCaDe relies on geometric consistency metrics derived from 2D and 3D object detections. Since object detections are already produced by standard autonomous driving perception stacks, this signal is available with essentially no additional sensing or inference cost.

\subsection{Decision Rules for Miscalibration Detection}

Beyond the monitored signal itself, existing methods differ in how calibration failures are detected. Levinson and Thrun~\cite{Levinson2013AutomaticOC} compare geometric consistency measures against fixed thresholds. Learning-based approaches formulate detection as a supervised binary classification problem, where each frame is independently classified as calibrated or miscalibrated~\cite{cross-modal-matching,tahiraj2025cal}.

While effective for large misalignments, frame-level decision rules suffer from several limitations. These approaches require supervised training or manually selected thresholds, do not accumulate evidence over time, and provide no explicit control over the false-alarm rate. In contrast, MiCaDe formulates calibration monitoring as a sequential change-detection problem, enabling training-free monitoring with a nominally controlled false-alarm rate.

\subsection{Extrinsic LiDAR-Camera Calibration}

Extrinsic calibration has been extensively studied as a parameter estimation problem. Classical methods rely on explicit geometric correspondences between LiDAR points and image features \cite{survey-calibration,pbacalib,intensity-calibration,sem-align}, while learning-based approaches directly regress calibration parameters through deep multimodal architectures \cite{iyer2018calibnet,calib-depth,yuan2020rggnet,ou2024iterative}. 

While calibration methods exploit cross-modal correspondences to estimate extrinsic parameters, MiCaDe addresses the complementary, operational problem of deciding \emph{when} such a correction should be invoked: even an accurate calibration procedure still needs to know when it should be invoked, since running it continuously is neither necessary nor safe, as it may overwrite a valid calibration. MiCaDe acts as a lightweight trigger, after which any compatible calibration module can be invoked to estimate the correction.
\section{Problem Formulation}
\label{sec:problem-formulation}

We consider a standard vision system equipped with a calibrated LiDAR-camera pair. Let $\Intrinsics \in \mathbb{R}^{3 \times 3}$ denote the known camera intrinsics, and let $\Rotation \in \mathbb{R}^{3 \times 3}$, $\vec{t} \in \mathbb{R}^{3}$ denote the extrinsic rotation and translation, respectively. Together they define the projection matrix $\ProjMatrix = \Intrinsics[\Rotation|\vec{t}] \in \mathbb{R}^{3 \times 4}$, which maps LiDAR coordinates to image coordinates.

Under nominal conditions, $\ProjMatrix$ is estimated during system setup and assumed fixed. However, mechanical perturbations, vibrations, or gradual thermal drift may alter the relative pose over time, introducing systematic inconsistencies between LiDAR and camera observations that accumulate and degrade downstream perception.

We assume to have access to synchronized camera--LiDAR streams $\{I_t\}_{t=0}^{T} \text{and} \{\mathcal{P}_t\}_{t=0}^{T}$, where $I_t$ denotes the RGB image and $\mathcal{P}_t$ the LiDAR point cloud at time $t$. We also assume access to one or more reference sequences $\mathcal{S}_{ref} = \{(I_t, \mathcal{P}_t)\}$ acquired under correct calibration, used to characterise the nominal behaviour of the system.

We further assume the vehicle is equipped with onboard 2D and 3D object detectors, already part of the perception stack, trained to recognize the same set of object classes so that the same car, pedestrian, or truck can be extracted from both modalities. At each time $t$, the LiDAR-based detector produces a set of $L$ 3D bounding boxes $\mathcal{D}^{3d}_t = \{(\bboxLidar{i}^t, s_i^t, \lambda_i^t)\}_{i=1}^{L}$, where $\bboxLidar{i} = (\vec{C}^{3d}_i, l_i, h_i, w_i, \yaw{i})^{\top}$ encodes the box center, size, and orientation, $s_i\in[0,1]$ its confidence score, and $\lambda_i$ its predicted class; the RGB-based detector similarly produces a set of $\ell$ 2D bounding boxes $\mathcal{D}^{2d}_t = \{(\bboxImg{j}^t, s_j^t, \lambda_j^t)\}_{j=1}^{\ell}$, centered in $\vec{c}_j^{2d}$.


\smallskip
\noindent\textbf{Miscalibration Detection.}
Let $\ProjMatrix_0$ denote the nominal projection matrix obtained during system setup. During operation, mechanical perturbations or gradual drift may change the effective calibration to an unknown projection matrix $\ProjMatrix_t$. Our goal is to detect the earliest time $\tau$ at which the calibration departs from its nominal value, i.e.,
\[
\ProjMatrix_t =
\begin{cases}
\ProjMatrix_0, & t < \tau,\\
\ProjMatrix_1 \neq \ProjMatrix_0, & t \geq \tau,
\end{cases}
\]
where $\ProjMatrix_1$ denotes the post-change projection matrix.
\section{Method: MiCaDe}
\label{sec:method}

\begin{figure*}[t]
\centering
\includegraphics[width=\textwidth]{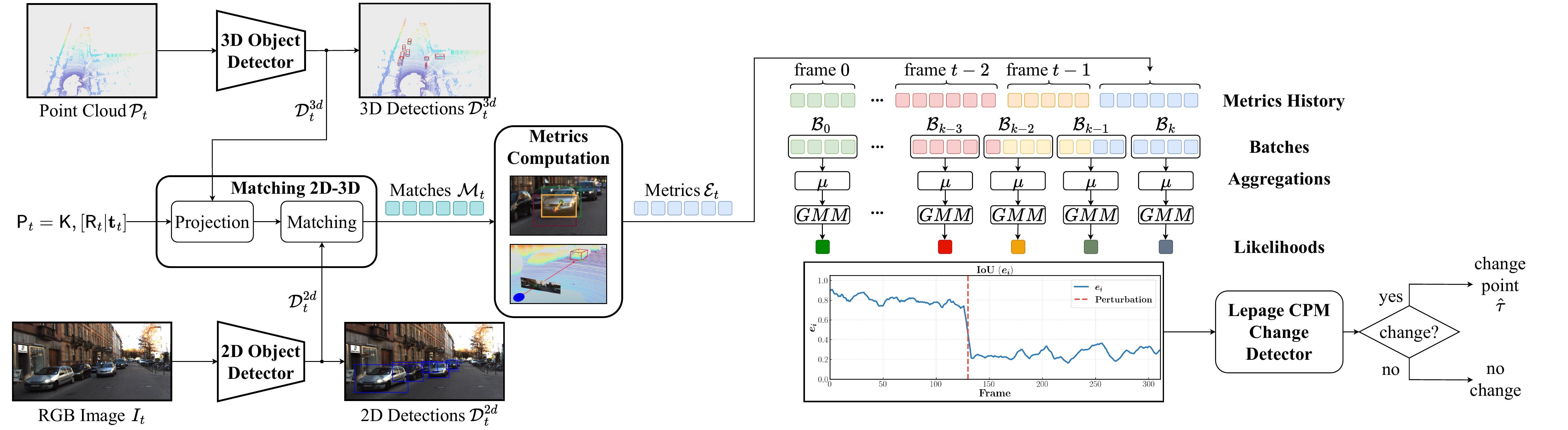}
\caption{\textbf{Overview of MiCaDe.}
At each timestep, 3D bounding boxes produced by a LiDAR detector are projected into the image plane and matched with 2D bounding boxes from an RGB detector.
For each matched pair $\vec{m}_k^t$, MiCaDe computes a geometric consistency vector $\vec{e}(\vec{m}_k^t)$ and appends it to a metrics history $\mathcal{H}$.
Metric vectors are grouped into fixed-size batches $\mathcal{B}_k$, aggregated into batch-level features $\vec{f}_k$, and scored against a reference GMM to obtain a likelihood stream $\{l_k\}$.
A Lepage CPM monitors this stream and raises an alarm at the estimated change point $\hat{\tau}$ when a persistent distributional shift is detected.}
\label{fig:main-figure}
\end{figure*}

MiCaDe, illustrated in Figure~\ref{fig:main-figure} and summarized in Algorithm~\ref{alg:complete-algorithm}, transforms each frame into a collection of geometric consistency indicators extracted from projected 3D bounding boxes matched with 2D detections. At each timestep $t$, using the nominal projection matrix $\ProjMatrix_0$, each 3D detection in $\mathcal{D}^{3d}_t$ is projected into the image plane and matched to 2D detections in $\mathcal{D}^{2d}_t$ via \emph{Matching} (line~\ref{ln:matching}), as detailed in Section~\ref{sec:late-fusion}, and a vector of geometric consistency metrics is computed (line~\ref{ln:per-match-metrics}), as in Section~\ref{subsec:metrics}.

Matched pairs are appended to a running buffer (line~\ref{ln:buffer}) and grouped into fixed-size batches (line~\ref{ln:batches}). Our intuition is that, under correct calibration, the projection of 3D bounding boxes fluctuates around their matched 2D counterparts with no systematic bias. When the relative camera-LiDAR pose drifts, instead, the projected boxes depart from the 2D detections in a consistent direction, inducing a distributional shift in the metric stream. We therefore adopt a sequential change-detection test (Section~\ref{sec:change-detection}) to monitor this stream and raise an alarm when a persistent shift is detected (line~\ref{ln:change-detection}). Since all operations are performed on a small set of bounding box detections, MiCaDe is suitable for continuous onboard monitoring with negligible computational overhead.

\begin{algorithm}[t]
\caption{MiCaDe}\label{alg:complete-algorithm}
\hspace*{\algorithmicindent} \textbf{Input}: Sequence of multimodal 
detections $(\mathcal{D}^{3d}_0, \mathcal{D}^{2d}_0), \dots, 
(\mathcal{D}^{3d}_T, \mathcal{D}^{2d}_T)$, initial calibration 
$\ProjMatrix_0 = \Intrinsics[\Rotation_0|\Translation_0]$, reference sequences $S_{ref}$ \\
\hspace*{\algorithmicindent} \textbf{Output}: Estimated change point 
$\hat{\tau}$
\begin{algorithmic}[1]
\Function{MiCaDe}{$\{\mathcal{D}^{3d}_t\}_{t=0}^T, \{\mathcal{D}^{2d}_t\}_{t=0}^T, \ProjMatrix_0, S_{ref}$}
\State $k \leftarrow 0$ \hfill \Comment{Batch index}
\State $\mathcal{H} \leftarrow \emptyset$ \hfill \Comment{Metrics history}
\State $\phi_0 \leftarrow \Call{FitGMM}{S_{ref}}$
\For{$t = 0, \dots, T$}
    \State $\mathcal{M}_t \leftarrow 
        \Call{Matching2D3D}{\mathcal{D}^{3d}_t, \mathcal{D}^{2d}_t, \ProjMatrix_0}$ \label{ln:matching}
    \For{$\vec{m} \in \mathcal{M}_t$}
        \State $\vec{e}(\vec{m}) \leftarrow \Call{ComputeMetrics}{\vec{m}}$ \label{ln:per-match-metrics}
        \State $\mathcal{H} \leftarrow \mathcal{H} \cup \{(\vec{m},\, \vec{e}(\vec{m}))\}$ \label{ln:buffer}
    \EndFor
    \State $\mathcal{B}_{k}, \dots, \mathcal{B}_{k+n_t} \leftarrow 
        \Call{GetNewBatches}{\mathcal{H}}$ \label{ln:batches}
    \State $\vec{f}_{k}, \dots, \vec{f}_{k+n_t} \leftarrow 
        \Call{Agg}{\mathcal{B}_{k}, \dots, 
        \mathcal{B}_{k+n_t}}$ \label{ln:aggregation}
    \State $l_{k}, \dots, l_{k+n_t} \leftarrow 
        \Call{GMM}{\vec{f}_{k}, \dots, \vec{f}_{k+n_t},\, 
        \phi_0}$ \label{ln:gmm}
    \State $k \leftarrow k + n_t$
    \State $\hat{k},\, \mathit{changed} \leftarrow 
        \Call{ChangeDetector}{\{l_{i}\}_{i=k}^{n_t}}$ 
        \label{ln:change-detection}
    \If{$\mathit{changed}$}
        \State \Return $\hat{\tau}$ \hfill 
            \Comment{Last timestamp of batch $\mathcal{B}_{\hat{k}}$}
    \EndIf
\EndFor
\State \Return $-1$ \hfill \Comment{No change detected}
\EndFunction
\end{algorithmic}
\end{algorithm}

\subsection{Cross-modal Correspondence Extraction}
\label{sec:late-fusion}

To match each LiDAR detection $\bboxLidar{i} \in \mathcal{D}^{3d}_t$ with at most one RGB detection $\bboxImg{j} \in \mathcal{D}^{2d}_t$, we follow the Bounding Box Matching procedure of LCF3D \cite{sgaravatti2026lcf3d}:
\begin{itemize}
    \item Each 3D bounding box $\bboxLidar{i}$ is projected into the image plane, obtaining the corresponding 2D box $\bboxGeneric{\text{proj}}_i$.
    \item A linear sum assignment problem is solved to find the one-to-one matching between all the projected boxes $\bboxGeneric{\text{proj}}_j$ and RGB detections $\bboxImg{j}$, maximizing their 2D IoU; considering only pairs sharing the same predicted class $\lambda_i = \lambda_j$ as candidates.
    \item Pairs whose IoU falls below a threshold $\rho = 0.1$ are discarded. We intentionally adopt a relatively low threshold, since the goal is not to guarantee highly accurate matches but rather to retain enough correspondences for monitoring. A higher threshold would reject many valid matches under moderate miscalibrations, precisely when detecting calibration drift is most important.
\end{itemize}

Since matching is itself performed on projected boxes, any miscalibration introduces inconsistencies between the 2D and 3D detections being matched, which could in principle give rise to wrong correspondences and reduce detection power. We verify in Section~\ref{sec:supp-matching-robustness} of the supplementary material, and report in Section~\ref{sec:experiments}, that matches remain reliable even under miscalibration, while the average number of valid matches per frame decreases, affecting detection delay rather than detection power.

Alternative robust alignment solutions as AutoAlign~\cite{chen2022autoalign} and GraphAlign~\cite{song2023graphalign} establish 2D-3D correspondences via a cross-modal feature interaction module and via graph matching, respectively. Both rely on learnable components trained end-to-end for detection, adding an overhead that conflicts with the training-free simplicity we target.

We denote by $\mathcal{M}_t =  \left\{ \vec{m}_k^t\right\}_{k= 1}^{N_t}$ the set of matched 3D–2D detections $\vec{m}_k^t= (\bboxLidar{i_k}^t, \bboxImg{j_k}^t)$ obtained at time step $t$, where the number of valid matches $N_t = |\mathcal{M}_t|$ depends on both scene content and calibration accuracy.

\subsection{Geometric Consistency Metrics}
\label{subsec:metrics}

\begin{figure}[t]
\centering
\includegraphics[width=\linewidth]{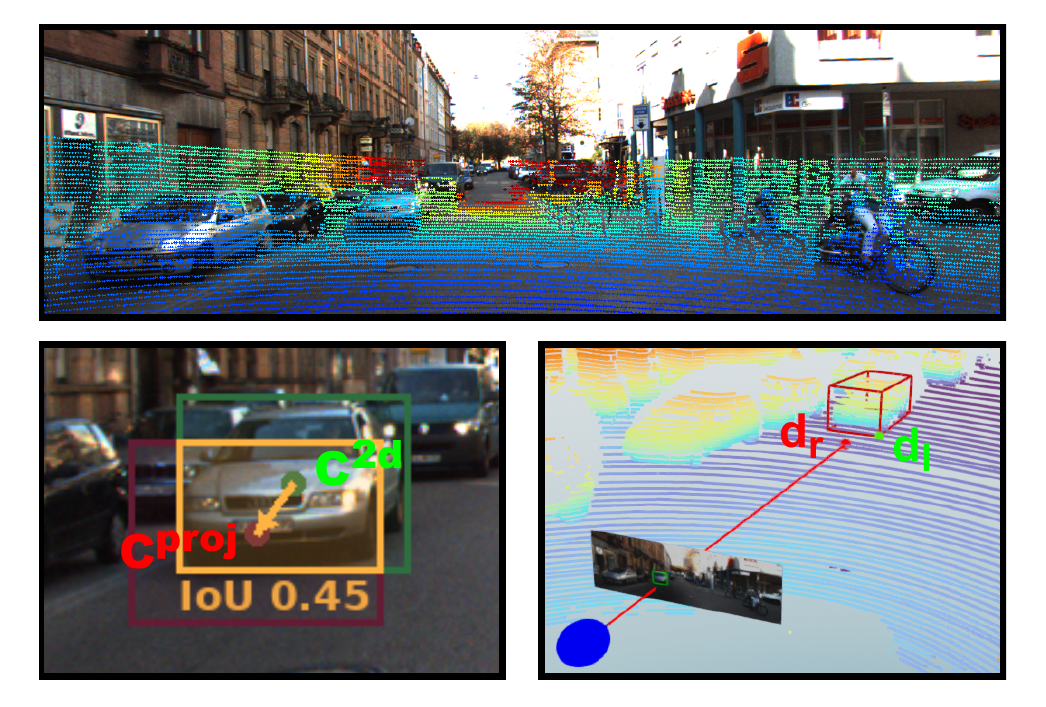}
\caption{(Top) A point cloud projected into the image with uncalibrated extrinsics. (Bottom left) IoU and center displacement between the projected 3D LiDAR bounding box and the corresponding 2D RGB bounding box. (Bottom right) The backprojection of the 2D bounding box produces an inaccurate depth estimate under miscalibration.}
\label{fig:metrics}
\end{figure}

For each match $\vec{m}_k^t = (\bboxLidar{i_k}^t,\bboxImg{j_k}^t)$, we derive a set of geometric quantities that characterize the consistency between the projected 3D detection and its corresponding 2D detection.
First, we define the center displacement vector
\begin{equation}
\vec{a}_k^t = \vec{c}^{2d}_{j_k}-\vec{c}^{\mathrm{proj}}_{i_k} \in \mathbb{R}^2,
\end{equation}
which measures the offset between the projected LiDAR box center and the image detection center. Under correct calibration, $\vec{a}_k^t$ fluctuates around $(0,0)$, while miscalibration induces a systematic bias in both magnitude and direction. We then define the displacement direction as $\alpha_k^t$, i.e. the angle of $\vec{a}_k^t$ with respect to the horizontal axis. Rather than using $\alpha_k^t$ directly, we encode the direction through its sine and cosine components. This avoids the discontinuity at $\pm\pi$, where two nearly identical directions correspond to numerically distant angle values.

Second, we compute the overlap between the two detections through their 2D Intersection-over-Union
\begin{equation}
e_i = \mathrm{IoU}\left(\bboxLidar{i_k}^{\mathrm{proj},t}, \bboxImg{j_k}^t \right).
\end{equation}
Finally, we estimate a coarse depth-consistency signal by backprojecting the bottom center of $\bboxImg{j_k}^t$ and intersecting it with the ground plane (assuming the bottom of $\bboxImg{j_k}^t$ lies on the ground), obtaining a depth estimate $d_r$, and comparing it with the LiDAR depth $d_l$ associated with $\bboxLidar{i_k}^t$ (Figure~\ref{fig:metrics}). These quantities are combined into the metric vector
\begin{equation}\label{eq:metrics-match}
\vec{e}(\vec{m}_k^t) = (e_x,e_y,e_i,e_a,e_{\sin},e_{\cos},e_b)^\top \in \mathbb{R}^{7},
\end{equation}
whose components are summarized in Table~\ref{tab:metrics}.

\begin{table}[t]
\centering
\caption{Per-match geometric consistency metrics.}
\label{tab:metrics}
\resizebox{\linewidth}{!}{
\begin{tabular}{lll}
\toprule
\textbf{Metric} & \textbf{Definition} & \textbf{Calibration cue} \\
\midrule

$x$-displacement $e_x$
&
$[\vec{a}_k^t]_x$
&
Horizontal projection offset
\\

$y$-displacement $e_y$
&
$[\vec{a}_k^t]_y$
&
Vertical projection offset
\\

Magnitude $e_a$
&
$\|\vec{a}_k^t\|_2$
&
Projection error magnitude
\\

Direction
$(e_{\sin},e_{\cos})$
&
$\sin(\alpha_k^t),\cos(\alpha_k^t)$
&
Direction of systematic drift
\\

IoU $e_i$
&
$\mathrm{IoU}(\bboxLidar{i_k}^{\mathrm{proj},t},
\bboxImg{j_k}^t)$
&
Overall 2D box overlap
\\

Depth error $e_b$
&
$\frac{|d_r-d_l|}{d_l}$
&
Coarse 3D consistency
\\

\bottomrule
\end{tabular}
}
\end{table}

\subsection{Calibration Monitoring}
\label{sec:change-detection}

\begin{figure}[t]
\centering
\includegraphics[width=\linewidth]{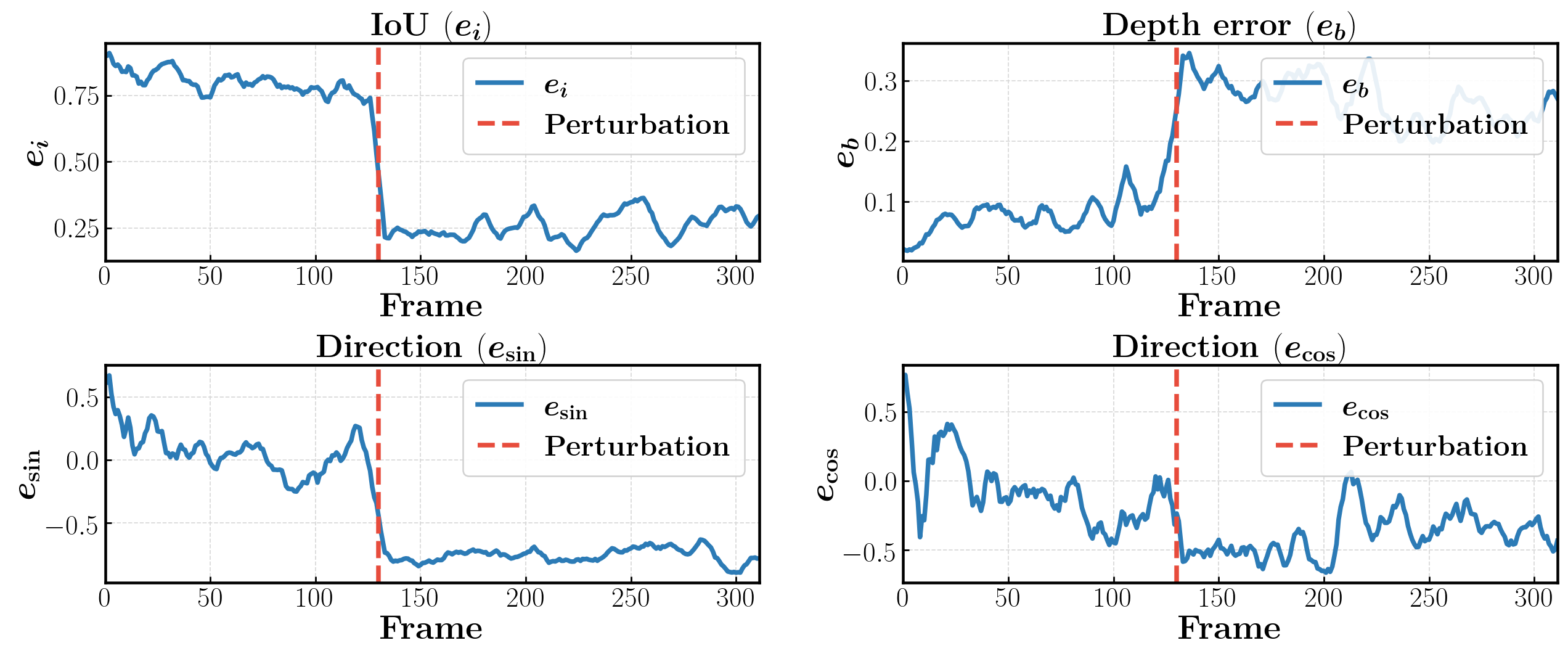}
\caption{\textbf{Example of monitored metrics evolution under calibration drift.}
A calibration perturbation is introduced after frame 50 (red dashed line). Metrics such as IoU ($e_i$), depth error ($e_b$), and directions ($e_sin$, $e_cos$) exhibit a clear distributional shift after the change, motivating the use of sequential change detection.}
\label{fig:changes}
\end{figure}

A drift in the extrinsic parameters induces a change in the distribution of the metric vectors $\vec{e}(\vec{m}_k^t)$ defined in \eqref{eq:metrics-match} (Fig.~\ref{fig:changes}). In particular, we expect $e_i$ to decrease, while $e_x$, $e_y$, $e_a$, and $e_b$ increase due to growing projection inaccuracies. Moreover, $e_{\sin}$ and $e_{\cos}$ transition from approximately zero-mean, high-variance variables to concentrated variables with non-zero mean. Rather than reacting to individual noisy detections, our objective is to identify a \emph{persistent} multivariate distributional shift, which naturally motivates the use of sequential change-detection techniques~\cite{ross2011nonparametric}.

Before processing the computed matches, we discard the unreliable ones. Specifically, a match is discarded if the projected 3D box falls for more than a fraction $\tau_t$ outside the image plane (truncated box), or if its 3D position lies farther than a distance $\tau_d$ from the sensor, since these induce a larger, calibration-independent localization noise.

\smallskip
\noindent\textbf{Batch aggregation.}
The number of valid matches $N_t$ varies across frames. Directly monitoring frame-level aggregates of the metrics in \eqref{eq:metrics-match} would violate the i.i.d. assumptions underlying change-detection tests, as their distribution would depend on $N_t$. To mitigate this issue, we accumulate metric vectors into a history buffer $\mathcal{H}$ and group them into fixed-size batches $\mathcal{B}_k = \{\vec m_{k_1}^{t_1}, \ldots, \vec m_{k_n}^{t_n}\}$, possibly spanning multiple consecutive frames.

The metrics are aggregated within each batch by computing the mean metric vector
\begin{equation}
\bar{\vec e}
=
\frac{1}{n}
\sum_{\vec m \in \mathcal B_k}
\vec e(\vec m),
\label{eq:batch-mean}
\end{equation}
and the variances of the directional components
\begin{equation}
\sigma^2_{\sin}
=
\frac{1}{n}
\sum_{\vec m \in \mathcal B_k}
\left(
e_{\sin}(\vec m)
-
\bar{e}_{\sin}
\right)^2,
\label{eq:batch-var}
\end{equation}
and analogously for $\sigma^2_{\cos}$.

The resulting feature vector $\vec{f}_k \in \mathbb R^9$ of the batch is
\begin{equation}
\vec f_k
=
(
\bar e_{x},
\bar e_{y},
\bar e_{i},
\bar e_{a},
\bar e_{\sin},
\bar e_{\cos},
\bar e_{b},
\sigma^2_{\sin},
\sigma^2_{\cos}
)^\top .
\label{eq:feature-vector}
\end{equation}

Under nominal operating conditions, the sequence $\{\vec f_k\}$ follows a multivariate distribution $\phi_0$ reflecting detector noise and scene variability.

\smallskip
\noindent\textbf{Likelihood Computation.}
We estimate the distribution$\phi_0$ of $\{\vec f_k\}$ extracted from one or more calibrated reference sequences $\mathcal S_{\mathrm{ref}}$ as a Gaussian Mixture Model (GMM). The GMM remains fixed throughout deployment. For each incoming batch, we compute the log-likelihood $l_k = \log p_{\phi_0}(\vec f_k)$ under the reference GMM.

\smallskip
\noindent\textbf{Change-point detection.} 
During operations, the stream of likelihoods $\{l_k\}$ is monitored using the non-parametric Change Point Model (CPM) of Ross \emph{et al.}~\cite{ross2011nonparametric}, based on the Lepage statistic. The CPM operates online and detects both changes in mean and variance of the monitored process. Moreover, it has two important features: (i) it provides control over the nominal Average Run Length ($ARL_0$), i.e., the expected time between false alarms, and (ii) it estimates, after each detection, the most likely change point $\hat k$, which is converted into the corresponding frame index $\hat \tau$. The decision thresholds corresponding to a target $ARL_0$ are pre-computed offline on synthetic sequences containing random noise realizations, following~\cite{ross2011nonparametric}. Unfortunately, ARL$_0$ control is exact only when the CPM inputs are i.i.d. realizations of $\phi_0$; as we show in Section~\ref{sec:experiments}, this assumption holds only approximately in our setting. Thus, the ARL$_0$ is intended as a hyperparameter that regulates the detection sensitivity.

\smallskip
\noindent\textbf{Computational complexity.}
MiCaDe operates exclusively on detections already produced by the perception stack and therefore introduces only a lightweight monitoring overhead. Its cost is dominated by the 2D-3D matching stage, a Hungarian assignment problem of complexity $O\!\left(\max(L_t,\ell_t)^3\right)$ in the number of LiDAR and image detections $L_t,\ell_t$; metric computation, batch aggregation, GMM evaluation, and CPM updates all operate on low-dimensional quantities and add negligible overhead. Overall, MiCaDe's cost scales with the number of detected objects rather than with the image or point-cloud size.

\section{Experiments}
\label{sec:experiments}

We test MiCaDe on three public autonomous driving datasets corrupted by synthetic perturbations, and on a real-world dataset with a physically induced calibration change.

\subsection{Implementation Details}

In our experiments, matches are filtered using a truncation threshold $\tau_t=0.2$ and a distance threshold $\tau_d=50$~m (Section~\ref{sec:change-detection}). Feature vectors extracted from $\mathcal{S}_{ref}$ are grouped using a batch size $n=15$ and used to fit a GMM, whose number of components is selected via the Bayesian Information Criterion. Incoming batches are scored using the fixed GMM and monitored online by the Lepage CPM, which uses $ARL_0=10000$ and $ARL_0=20000$. Further implementation and reproducibility details are provided in Section~\ref{sec:supp-implementation} of the supplementary material.

\subsection{Datasets and Synthetic Perturbations}
\label{sec:datasets}

\smallskip\noindent\textbf{Synthetic Datasets.}\label{sec:perturbation}
We evaluate MiCaDe on three public autonomous driving datasets: KITTI Raw~\cite{geiger2013vision}, ZOD~\cite{zod} and ONCE~\cite{once}. As object detectors, we adopt PV-RCNN~\cite{shi2020pv} and Faster R-CNN~\cite{ren2015faster} on KITTI Raw, PointPillars~\cite{lang2019pointpillars} and Faster R-CNN on ZOD, and PointPillars and DDQ~\cite{zhang2023dense} on ONCE. KITTI and ZOD run at 10~Hz while ONCE provides frames at 2~Hz, which we use to convert delay to seconds. We use as reference set $\mathcal{S}_{ref}$ a subset of the detectors' training sequences, and evaluate on the remaining sequences. Further details on dataset splits are provided in Section~\ref{sec:supp-implementation}.

We inject synthetic calibration drifts into all the sequences following two protocols, \emph{abrupt changes} and \emph{gradual changes}; for each sequence, we generate 50 perturbed variants, randomly sampling the perturbation parameters and the change point $\tau$, and report results averaged over all variants and sequences. We select $\tau$ at least 100 frames after the sequence start and 100 frames before the end of the sequence, and apply a rigid transformation
\[
\Delta \vec{T} =
\begin{bmatrix}
\Delta \vec{R} & \Delta \vec{t} \\
0 & 1
\end{bmatrix}
\]
to all LiDAR observations acquired after $\tau$, leaving the camera stream unchanged. For \emph{abrupt changes}, $\Delta \vec{R}$ is generated by independently sampling roll, pitch, and yaw deltas from $[-0.05,\,0.05]\,\mathrm{rad}$, and $\Delta \vec{t}$ from $[-0.2,\,0.2]\,\mathrm{m}$, independently for each variant, applied at once at $\tau$. For \emph{gradual changes}, $\Delta \vec{T}$ is instead ramped up linearly from identity to a fixed magnitude ($3^\circ$ on all rotation angles, $0.2$~m on all translation components) over 100 frames starting at $\tau$, remaining constant thereafter.

\smallskip\noindent\textbf{Real-World Dataset.}
We evaluate MiCaDe on a one-hour sequence from a platform with three RGB cameras and a 128-beam LiDAR (Section~\ref{sec:real-world}). Unlike existing benchmarks, our real-world dataset contains a \emph{physically induced} calibration change, capturing additional sources of variability.
A known calibration drift is introduced by rotating one camera by approximately $3^\circ$, resulting in 2,151 calibrated and 5,234 miscalibrated frames. The first 1,000 calibrated frames are used for $\mathcal{S}_{ref}$, while the remainder is processed online. Detectors trained on ZOD are used in this setting. The dataset will be publicly released upon acceptance.

\subsection{Competitors}

We compare against Tahiraj \emph{et al.}~\cite{tahiraj2025cal}, trained using MiCaDe's reference split and the protocol of~\cite{tahiraj2025cal}. During evaluation, it flags each frame independently as calibrated or miscalibrated, using both the nominal classification threshold (0.5) and also an adjusted threshold on clean sequences to match MiCaDe's empirical $ARL_0=10000$, enabling a comparison at matched false-alarm rates. Levinson and Thrun~\cite{Levinson2013AutomaticOC} and Chen~\emph{et al.}~\cite{cross-modal-matching} are also related, but lack an official implementation and could not be included.

\subsection{Evaluation Metrics}

We evaluate MiCaDe on clean sequences and on sequences with an abrupt or gradual calibration change. For each perturbed sequence, we consider the first alarm raised at time $T$: if $T > \tau$, the change is correctly detected with delay $|T-\tau|$; if $T < \tau$, it is a false positive (FP), excluded from the delay computation; if no detection is raised, it is a false negative (FN), which should not occur with sufficiently long sequences. On clean sequences we compute the empirical $ARL_0$ (average time, in seconds, between false alarms) and FP/h (false alarms per hour). On sequences with abrupt or gradual changes: the False Positive Rate (FPR) and False Negative Rate (FNR), as above, together with the mean Detection Delay in seconds. On gradual changes only we also report the angle, in degrees, reached by the ramp at the time of detection.
In the supplementary material, we report the delay distribution's density and cumulative distribution (Section~\ref{sec:supp-delay-distribution}).
We additionally report the AUC, the area under the ROC curve, treating the calibrated and miscalibrated portions of a sequence as the two classes and the monitored signal (log-likelihood for MiCaDe, classification score for Tahiraj \emph{et al.}) as the score, computed on sequences with abrupt changes, which measures a signal's raw discriminative power independently of the monitoring scheme.

\begin{table*}[t]
\centering
\small
\resizebox{\textwidth}{!}{%
\begin{tabular}{lllccccccccccccc}
\toprule
\multirow{2}{*}{Dataset} & \multirow{2}{*}{Method} & \multirow{2}{*}{Threshold} & \multicolumn{3}{c}{Clean sequences} & \multicolumn{3}{c}{Abrupt changes} & \multicolumn{4}{c}{Gradual changes} & \multirow{2}{*}{Speed [ms] $\downarrow$} & \multirow{2}{*}{AUC $\uparrow$} \\
\cmidrule(lr){4-6} \cmidrule(lr){7-9} \cmidrule(lr){10-13}
& & & ARL$_0$ [s] $\uparrow$ & ARL$_0$ [b] $\uparrow$ & FP/h $\downarrow$ & FNR $\downarrow$ & FPR $\downarrow$ & Delay [s]  $\downarrow$ & FNR $\downarrow$ & FPR$\downarrow$ & Delay [s] $\downarrow$ & Angle [deg] $\downarrow$ & & \\
\midrule
\multirow{4}{*}{KITTI}
& MiCaDe & $ARL_0=10000$ & 617.4 & 398.5 & 5.83 & 0.088 & 0.001 & 8.78 & 0.044 & 0.043 & 8.47 & 1.865 & \multirow{2}{*}{2} & \multirow{2}{*}{0.9976}\\
& MiCaDe & $ARL_0=20000$ & - & - & - & 0.161 & 0.000 & 8.92 & 0.044 & 0.000 & 8.77 & 1.943 & & \\
\cmidrule(lr){2-15}
& Tahiraj \emph{et al.} & .5 & 6.00 & - & 597.6 & 0.000 & 0.651 & 0.40 & 0.000 & 0.594 & 3.20 & 0.992 & \multirow{2}{*}{12*} & \multirow{2}{*}{0.9969} \\
& Tahiraj \emph{et al.} (adj.) & .999998 & 595.2 & - & 6.48 & 0.024 & 0.012 & 0.60 & 0.031 & 0.031 & 4.40 & 1.364 & & \\
\midrule
\multirow{4}{*}{ZOD}
& MiCaDe & $ARL_0=10000$ & 941.8 & 1041.4 & 3.82 & 0.000 & 0.000 & 4.66 & 0.000 & 0.000 & 21.43 & 1.148 & \multirow{2}{*}{2} & \multirow{2}{*}{0.9965} \\
& MiCaDe & $ARL_0=20000$ & 1177.3 & 1666.2 & 3.06 & 0.006 & 0.000 & 5.66 & 0.000 & 0.000 & 24.50 & 1.348 & & \\
\cmidrule(lr){2-15}
& Tahiraj \emph{et al.} & .5 & 38.5 & - & 93.41 & 0.000 & 0.417 & 0.40 & 0.000 & 0.334 & 3.37 & 0.185 & \multirow{2}{*}{14*} & \multirow{2}{*}{1.000} \\
& Tahiraj \emph{et al.} (adj.) & .9974 & 932.4 & - & 3.79 & 0.865 & 0.000 & 0.59 & 1.000 & 0.000 & - & - & & \\
\midrule
\multirow{4}{*}{ONCE}
& MiCaDe & $ARL_0=10000$ & 4523.5 & 1332.0 & 0.80 & 0.070 & 0.002 & 56.48 & 0.105 & 0.053 & 84.29 & 1.806 & \multirow{2}{*}{2} & \multirow{2}{*}{0.9989}\\
& MiCaDe & $ARL_0=20000$ & 8750.0 & 2664.0 & 0.41 & 0.147 & 0.000 & 68.22 & 0.105 & 0.000 & 134.10 & 2.822 & & \\
\cmidrule(lr){2-15}
& Tahiraj \emph{et al.} & .5 & 902.3 & - & 3.99 & 0.000 & 0.373 & 1.10 & 0.000 & 0.210 & 2.5 & 0.149 & \multirow{2}{*}{11*} & \multirow{2}{*}{1.000} \\
& Tahiraj \emph{et al.} (adj.) & .8278 & 4060.4 & - & 0.89 & 0.000 & 0.062 & 3.00 & 0.000 & 0.053 & 3.0 & 0.173 & & \\
\bottomrule
\end{tabular}%
}
\caption{Results on clean sequences, abrupt calibration changes, and gradual calibration changes. ARL$_0$ on clean sequences is reported both in seconds [s] and in batches [b]. Speed and AUC are reported once per method, as they do not depend on the ARL$_0$/threshold operating point. AUC measures the raw discriminative power of each signal on the sequences with abrupt changes, independently of the thresholds used. ``Tahiraj (adj.)'' denotes the threshold adjusted to match MiCaDe's empirical $ARL_0=10000$ on clean sequences. Values with an asterisk (*) denote GPU computations.}
\label{table:main-results-expanded}
\end{table*}

\subsection{Main Results}

Table~\ref{table:main-results-expanded} reports results on clean sequences and on sequences with abrupt and gradual calibration changes. We omit $ARL_0=1000$: assuming 2s per batch, it would still allow at least 2 FP/h if respected, too frequent for our target regime. On clean sequences, MiCaDe's empirical $ARL_0$ falls short of the nominal target used to calibrate the Lepage threshold on every dataset, whether measured in seconds or directly in batches. Part of this gap is expected, since the target is nominally defined on the batch index; the persisting shortfall in batch count indicates that the monitored batch-level features are not fully stationary. Raising the operating point from $ARL_0=10000$ to $ARL_0=20000$ substantially reduces the false-alarm rate at a modest cost in detection delay, and we adopt it as our recommended operating point; reliably estimating even rarer false-alarm rates would require longer clean sequences than available here, as shown by KITTI at $ARL_0=20000$, where no false alarm is observed in clean sequences as these are too short.

At its nominal threshold ($0.5$), Tahiraj \emph{et al.} yields impractically high FPR ($0.651$/$0.417$/$0.373$ on KITTI/ZOD/ONCE); matching MiCaDe's empirical $ARL_0$ instead pushes its threshold to extreme values ($0.999998$, $0.9974$). Even so, MiCaDe achieves lower FPR on KITTI/ONCE and far lower FNR on ZOD ($0.006$ vs.\ $0.865$), at comparable AUC ($\geq 0.99$ everywhere): our geometry-based indicators match Tahiraj's learned representation without supervised training or GPU inference.

The performance gap across datasets can be partly explained by sensing setup: KITTI is tightly synchronized with favourable geometry, whereas ZOD has only approximate synchronization and ONCE has a larger camera--LiDAR displacement, increasing projection variability. MiCaDe's FPR stays within $0.000$--$0.002$ across all three at $ARL_0=10000$; ONCE's larger displacement instead shows up in detection delay, the longest overall ($56.48$s vs.\ $8.78$s KITTI, $4.66$s ZOD). Delay itself is strongly skewed towards small values, as most changes are detected quickly while a smaller subset of harder sequences pulls the mean up substantially (e.g., median $7.61$~s vs.\ $90$th percentile $19.11$~s on KITTI Raw; Section~\ref{sec:supp-delay-distribution} reports the full distribution). Qualitative examples are provided in Section~\ref{sec:supp-qualitative}.

\smallskip\noindent\textbf{Sensitivity to individual miscalibration parameters.}
Unlike the randomized protocol of Section~\ref{sec:perturbation}, here we apply a single, deterministic perturbation (one axis, one magnitude) per sequence, evaluating each degree of freedom independently at three fixed magnitudes: $0.0125$, $0.025$, $0.05$~rad for rotations; $0.05$, $0.1$, $0.2$~m for translations (per-magnitude results are in Section~\ref{sec:supp-magnitude}). The perturbation is always injected at frame 100, isolating delay attributable to the perturbed DOF from change-point variability. Table~\ref{table:ablation-axes} reports FNR, FPR, and delay averaged across magnitudes on ZOD. Rotational drifts are detected faster and more reliably than translational ones, with $t_y$ the hardest axis due to its weaker influence on the monitored geometric cues compared to the systematic horizontal displacement induced by roll or yaw.

\begin{table}[t]
\centering
\begin{tabular}{lcccc}
\toprule
DOF & FNR$\downarrow$ & FPR$\downarrow$ & Delay [s] $\downarrow$ & AUC$\uparrow$ \\
\midrule
Roll   & 0.000 & 0.000 & 8.3 & 0.991 \\
Yaw    & 0.000 & 0.000 & 7.3 & 0.962\\
Pitch  & 0.067 & 0.000 & 21.9 & 0.749 \\
$t_x$  & 0.100 & 0.000 & 45.3 & 0.617 \\
$t_y$  & 0.267 & 0.000 & 91.2 & 0.482 \\
$t_z$  & 0.133 & 0.000 & 36.5 & 0.692 \\
\bottomrule
\end{tabular}
\caption{Sensitivity to individual miscalibration parameters on ZOD (ARL$_0$=10000). Rotational drifts are detected faster than translational ones.}
\label{table:ablation-axes}
\end{table}

\begin{figure*}[t]
\centering
\includegraphics[width=0.9\linewidth]{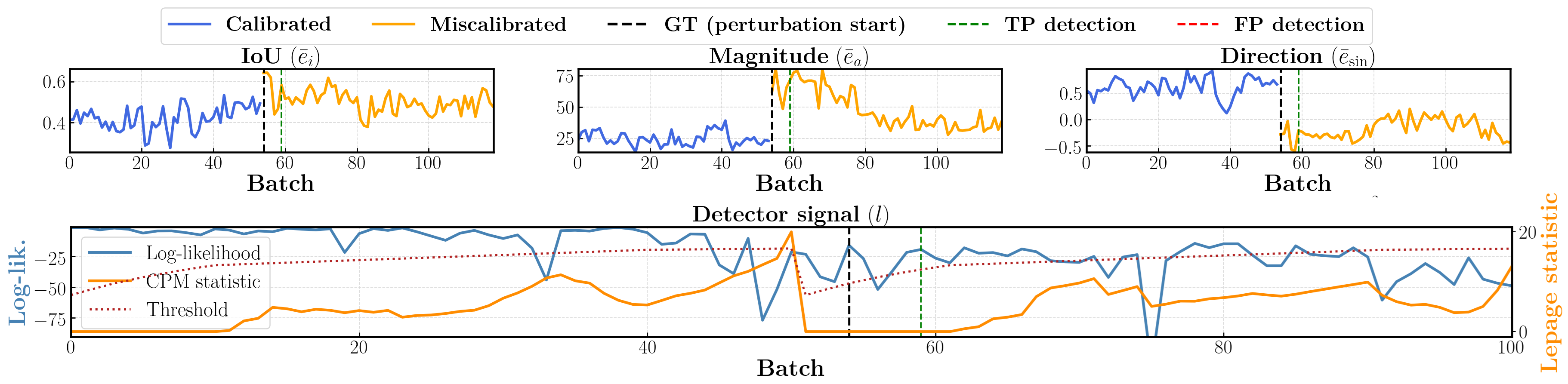}
\caption{
Evolution of three representative monitored metrics, IoU ($e_i$), displacement magnitude ($e_a$), and direction ($e_{\sin}$), together with the detector signal on the real-world sequence.
The black dashed line marks the manually induced calibration change, while the green line indicates the detection raised by MiCaDe.
The bottom panel reports the log-likelihood stream and the corresponding CPM statistic, which crosses the decision threshold shortly after the perturbation, correctly identifying the calibration drift.
The full set of monitored metrics is reported in Section~\ref{sec:dataset-supplementary} of the supplementary material.
}
\label{fig:metrics-aida}
\end{figure*}

\smallskip\noindent\textbf{Contribution of individual metrics.}
Table~\ref{table:ablation-metrics} reports the contribution of different feature subsets on ZOD, evaluated on the same perturbed sequences used in Table~\ref{table:main-results-expanded}. Among the individual groups, IoU ($e_i$) and the center-displacement pair ($e_x,e_y$) achieve the lowest FNR ($0.063$ and $0.051$) and the highest AUC ($0.972$ and $0.976$), while direction alone attains the lowest FPR ($0.000$) at the cost of a higher FNR ($0.165$); the depth-consistency pair ($e_a,e_b$) is weakest overall, with the highest FPR ($0.114$) and longest delay ($17.09$s). The directional variances alone yield the lowest standalone AUC ($0.845$). Overall, the full feature vector achieves the best FPR ($0.000$), delay ($4.66$s), and AUC ($0.997$), without any false negatives, confirming that variance terms complement the mean-based metrics.

\begin{table}[t]
\centering
\resizebox{\linewidth}{!}{%
\begin{tabular}{lccccccccccccc}
\toprule
\multirow{2}{*}{Configuration}
  & \multicolumn{9}{c}{Metrics used}
  & \multirow{2}{*}{FNR $\downarrow$}
  & \multirow{2}{*}{FPR $\downarrow$}
  & \multirow{2}{*}{Delay [s] $\downarrow$}
  & \multirow{2}{*}{AUC $\uparrow$} \\
\cmidrule(lr){2-10}
  & $e_x$ & $e_y$ & $e_i$ & $e_a$ & $e_b$ & $e_{\sin}$ & $e_{\cos}$ & $\sigma^2_{\sin}$ & $\sigma^2_{\cos}$ & & & \\
\midrule
$e_i$ only & & & \checkmark & & & & & & & 0.063 & 0.080 & 5.32 & 0.972 \\
$e_x, e_y$ only & \checkmark & \checkmark & & & & & & & & 0.051 & 0.040 & 8.55 & 0.976 \\
$e_a, e_b$ only & & & & \checkmark & \checkmark & & & & & 0.222 & 0.114 & 17.09 & 0.909 \\
Direction only & & & & & & \checkmark & \checkmark & & & 0.165 & 0.000 & 7.52 & 0.949 \\
Variance only & & & & & & & & \checkmark & \checkmark & 0.290 & 0.108 & 10.09 & 0.845 \\
No variance & \checkmark & \checkmark & \checkmark & \checkmark & \checkmark & \checkmark & \checkmark & & & 0.040 & 0.000 & 5.44 & 0.993 \\
Full (ours) & \checkmark & \checkmark & \checkmark & \checkmark & \checkmark & \checkmark & \checkmark & \checkmark & \checkmark & \textbf{0.000} & \textbf{0.000} & \textbf{4.66} & \textbf{0.997} \\
\bottomrule
\end{tabular}
}
\caption{Ablation study on the feature set used by MiCaDe on ZOD (ARL$_0$=10000). Each row indicates the components of $\vec{f}_k \in \mathbb{R}^9$ included in the change-detection pipeline.}
\label{table:ablation-metrics}
\end{table}

\smallskip\noindent\textbf{Robustness of the detector.}
We further verify that MiCaDe's alarms are not confused with unrelated sources of variability. 2D--3D matching remains reliable under miscalibration (Section~\ref{sec:supp-matching-robustness}): accuracy drops marginally, from $92\%$ to $89\%$, and the average number of matches per frame decreases (from 5 to 3), affecting delay rather than detection power. Neither scene changes (Section~\ref{sec:supp-scene-changes}) nor severe LiDAR degradation (Section~\ref{sec:supp-degradation}, e.g., a missing $90^\circ$ sector) trigger false alarms at $ARL_0=10000$, and a genuine miscalibration on top of a degraded stream is still detected.

\subsection{Real-World Case Study}
\label{sec:real-world}

Figure~\ref{fig:metrics-aida} shows the evolution of three of the monitored metrics, IoU, displacement magnitude, and direction, together with the detector output; the remaining metrics behave analogously and are reported in full in the supplementary material (Figure~\ref{fig:metrics-aida-full}). Following the induced perturbation, all monitored metrics exhibit a clear distributional shift, which propagates to the likelihood stream and causes the CPM statistic to cross the decision threshold after 34 frames, corresponding to only 5 batches in this setting, with no false alarms observed throughout the sequence. These results confirm that MiCaDe successfully distinguishes persistent calibration drift from normal variability, and that the behaviour observed under synthetic perturbations transfers to real operating conditions.

\section{Conclusions}
\label{sec:conclusions}

We presented MiCaDe, a lightweight framework for online LiDAR--camera calibration monitoring based on sequential change detection. By monitoring geometric consistency metrics extracted from standard 2D and 3D object detections, MiCaDe detects persistent calibration drift without supervised training or modifications to the perception stack.
Experiments on KITTI Raw, ZOD, ONCE, and a real-world dataset with physically induced miscalibration demonstrate reliable detection across different sensors and driving conditions, while classifier-based baselines require impractical thresholds to achieve MiCaDe's false-alarm rate.

A current limitation is the reliance on successful 3D--2D matching, which may fail in sparse scenes or under severe calibration errors. A second limitation, exposed by the gap between the nominal and empirical $ARL_0$ observed in Section~\ref{sec:experiments}, is that the monitored batch-level features do not fully satisfy the i.i.d.\ assumption underlying the CPM, so its false-alarm control holds only approximately. Future work will investigate alternative change-detection schemes more robust to non-stationary inputs, the integration of automatic recalibration modules triggered by MiCaDe, and whether learned alignment strategies such as AutoAlign~\cite{chen2022autoalign} or GraphAlign~\cite{song2023graphalign} can further improve detection power, at the cost of the training-free simplicity targeted in this work.
\section*{Acknowledgements}
This paper is supported by  FAIR  (NextGenerationEU program, PNRR-PE-AI scheme, M4C2, Investment 1.3, Line on Artificial Intelligence). We acknowledge the EuroHPC Joint Undertaking for awarding this project access to the EuroHPC supercomputer Karolina hosted by IT4Innovations, Czech Republic through an EuroHPC Development Access call.
{
    \small
    \bibliographystyle{ieeenat_fullname}
    \bibliography{main}
}

\clearpage
\setcounter{page}{1}
\maketitlesupplementary

This supplementary material provides additional details supporting the main paper. We first describe the implementation of MiCaDe and the real-world dataset used for evaluation. We then report extended experimental analyses, including sensitivity to perturbation magnitude, detection-delay distributions, and qualitative examples of both true-positive and false-negative cases.

\appendix

\section{MiCaDe: Implementation Details}
\label{sec:supp-implementation}

MiCaDe implementation for the experiments described in Section \ref{sec:experiments} is lightweight and designed for online deployment. For each dataset, the 2D and 3D object detectors are trained on the corresponding training split, while a subset of calibrated sequences is reserved to construct the reference set $\mathcal{S}_{ref}$ used to estimate the nominal behaviour of the monitored features. The detector architectures adopted for each dataset are reported in Section~\ref{sec:datasets} of the main paper. We use MMDetection as a framework for training the 2D object detector and MMDetection3D for training the 3D object detector.

Feature vectors extracted from $\mathcal{S}_{ref}$ are accumulated into batches of $n=15$ matched detections and used to fit a Gaussian Mixture Model (GMM) through the Expectation-Maximization (EM) algorithm. The number of Gaussian components is selected using the Bayesian Information Criterion (BIC), resulting in 4 components for KITTI Raw and 8 components for both ZOD and ONCE. The resulting GMM remains fixed during deployment.

At runtime, each incoming batch is evaluated under the reference GMM to obtain a log-likelihood score. The resulting likelihood stream is monitored online by the Lepage Change Point Model (CPM), configured with $\mathrm{ARL}_0=10000$ or $\mathrm{ARL}_0=20000$, which automatically determines the detection threshold corresponding to the desired average run length under nominal conditions.

Unless otherwise stated, the 3D--2D matching stage uses an IoU threshold of $0.1$, which was found to provide a good trade-off between retaining sufficient correspondences under moderate miscalibration and rejecting incorrect matches. Matches are additionally filtered using a truncation threshold $\tau_t=0.2$ and a distance threshold $\tau_d=50$~m (Section~\ref{sec:change-detection}), discarding detections whose localization noise is expected to be large independently of calibration. All monitoring operations are implemented on the CPU and involve only geometric computations and GMM likelihood evaluations over low-dimensional feature vectors, resulting in negligible computational overhead with respect to the perception pipeline.

\smallskip
\noindent\textbf{Dataset details.} KITTI Raw~\cite{geiger2013vision} provides 70 synchronized LiDAR--camera sequences acquired in urban, rural, and highway environments. Evaluation uses the 36 sequences with more than 350 frames; the remaining sequences are used for detector training and to construct $\mathcal{S}_{ref}$. ZOD~\cite{zod} provides both annotated single frames (ZOD Frames, used for detector training) and long driving sequences (ZOD Drives); ZOD Drives are further split into training drives, used to construct $\mathcal{S}_{ref}$, and validation drives, used for evaluation. ONCE~\cite{once} contains LiDAR--camera driving sequences captured under diverse environmental conditions: detectors are trained on the training split, which also estimates the reference distribution $\phi_0$, while evaluation is performed on the validation split.

\section{Additional Results}

\subsection{Sensitivity to Perturbation Magnitude}
\label{sec:supp-magnitude}

Table~\ref{tab:supp-ablation-magnitude} reports the complete results for the perturbation-magnitude study summarized in the main paper. While Table~\ref{table:ablation-axes} reports FNR, FPR, and mean detection delay aggregated across perturbation magnitudes, here we analyze each perturbation level individually.

For each degree of freedom, we perturb only one calibration parameter at a time. Rotational perturbations are evaluated at $0.0125$, $0.025$, and $0.05$~rad, whereas translational perturbations are evaluated at $0.05$, $0.1$, and $0.2$~m. For every configuration, we generate ten perturbed sequences by injecting the miscalibration at a random point in the sequence and evaluate MiCaDe using the protocol described in the main paper.

Overall, larger perturbations tend to be detected more quickly, since they induce greater changes in the monitored geometric consistency metrics and therefore require fewer observations before the sequential detector accumulates sufficient statistical evidence. This trend is particularly evident for pitch, $t_x$, and $t_z$, whose mean detection delay decreases substantially as the perturbation magnitude increases.

The relationship is, however, not strictly monotonic. For example, roll exhibits a slightly lower delay at $0.025$~rad than at $0.05$~rad, while the delay associated with the vertical translation $t_y$ remains consistently high across all magnitudes. These fluctuations are expected because the detection delay depends not only on the perturbation magnitude, but also on scene content, the number and spatial distribution of matched objects, and the stochastic nature of the sequential change-detection process. Nevertheless, the overall trend confirms that MiCaDe becomes increasingly responsive as the severity of the calibration drift grows.

\begin{table}[t]
\centering
\resizebox{\linewidth}{!}{
\begin{tabular}{llcccc}
\toprule
DOF & Magnitude & FNR$\downarrow$ & FPR$\downarrow$ & Delay [s]$\downarrow$ & AUC$\uparrow$ \\
\midrule
Roll  & 0.0125 rad & 0.000 & 0.000 & 10.43 & 0.974 \\
      & 0.0250 rad & 0.000 & 0.000 & 6.31 & 0.999 \\
      & 0.0500 rad & 0.000 & 0.000 & 8.28 & 1.000 \\
\midrule
Yaw   & 0.0125 rad & 0.000 & 0.000 & 11.72 & 0.920 \\
      & 0.0250 rad & 0.000 & 0.000 & 5.66 & 0.972 \\
      & 0.0500 rad & 0.000 & 0.000 & 4.58 & 0.994 \\
\midrule
Pitch & 0.0125 rad & 0.100 & 0.000 & 51.57 & 0.551 \\
      & 0.0250 rad & 0.100 & 0.000 & 7.12 & 0.749 \\
      & 0.0500 rad & 0.000 & 0.000 & 6.91 & 0.947 \\
\midrule
$t_x$ & 0.05 m & 0.200 & 0.000 & 63.47 & 0.516 \\
      & 0.10 m & 0.100 & 0.000 & 58.78 & 0.559 \\
      & 0.20 m & 0.000 & 0.000 & 13.53 & 0.775 \\
\midrule
$t_y$ & 0.05 m & 0.400 & 0.000 & 112.85 & 0.491 \\
      & 0.10 m & 0.200 & 0.000 & 97.61 & 0.472 \\
      & 0.20 m & 0.200 & 0.000 & 63.13 & 0.484 \\
\midrule
$t_z$ & 0.05 m & 0.300 & 0.000 & 52.32 & 0.537 \\
      & 0.10 m & 0.100 & 0.000 & 50.61 & 0.673 \\
      & 0.20 m & 0.000 & 0.000 & 6.55 & 0.865 \\
\bottomrule
\end{tabular}
}
\caption{FNR, FPR, and mean detection delay for each perturbation magnitude. Results are reported on ZOD over 10 perturbation trials per configuration.}
\label{tab:supp-ablation-magnitude}
\end{table}

\subsection{Distribution of Detection Delays}
\label{sec:supp-delay-distribution}

In the main paper, detection delay is summarized using the mean delay over true-positive detections. However, the delay distribution is not symmetric: most calibration changes are detected quickly, while a smaller number of sequences require substantially longer observation windows before enough statistical evidence is accumulated. To better characterize this behaviour, Figure~\ref{fig:supp-delay-distribution} reports the empirical distribution of detection delays on KITTI Raw.

The histogram shows a strongly right-skewed distribution. The median delay is $7.61$~s, meaning that half of the detected calibration changes are identified within a short temporal window. At the same time, the $90$th percentile is $19.11$~s, indicating the presence of a long tail of more challenging cases extending up to several tens of seconds. These late detections typically correspond to sequences with fewer reliable 3D--2D correspondences, weaker projection shifts, or higher scene variability, where the sequential detector needs more batches before crossing the decision threshold.

The empirical cumulative distribution function further confirms this behaviour: the detection probability increases rapidly in the first seconds after the perturbation and then saturates more slowly due to the long-tail cases. This illustrates why a single mean-delay figure does not fully capture this behavior, and highlights the importance of reporting delay statistics beyond the average.

\begin{figure}[t]
\centering
\includegraphics[width=\linewidth]{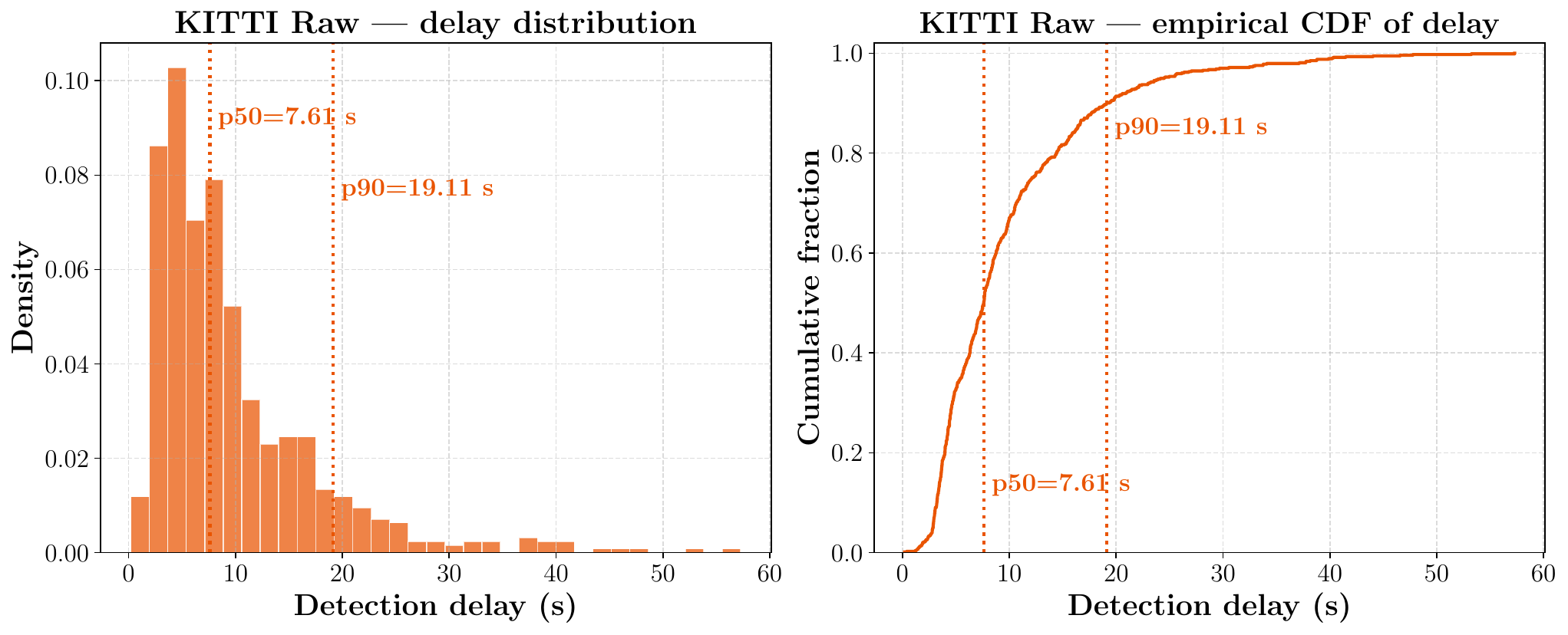}
\caption{
Distribution of detection delays on KITTI Raw. The left plot shows the delay histogram, while the right plot reports the empirical cumulative distribution function. Dotted vertical lines indicate the median delay ($p50=7.61$~s) and the $90$th percentile delay ($p90=19.11$~s).
}
\label{fig:supp-delay-distribution}
\end{figure}

\subsection{Robustness to Scene Changes}
\label{sec:supp-scene-changes}

We test whether a scene change, rather than a genuine calibration drift, can itself be mistaken for a persistent distributional shift. For each dataset, we concatenate pairs of clean sequences (A, B) and treat the concatenation point as a \emph{pseudo change point}: calibration is correct on both sides, so an alarm attributed to it would indicate confusion between scene change and calibration drift. For each pair, $\phi_0$ is refit excluding both A and B, and MiCaDe runs with the standard operating point ($ARL_0=10000$). As a dataset-agnostic proxy for scene difference, we use object density (average frames needed to fill a batch), forming \emph{heterogeneous} dense/sparse pairs against \emph{homogeneous} control pairs of matching density, drawn from the same pool of sequences; where richer metadata is available, we also pair sequences by country/daypart (ZOD) and weather/daypart (ONCE, limited to very few pairs). The primary metric is the \emph{junction-FPR}: the fraction of pairs with an alarm within $\pm 15$ batches of the junction. FP/h, the global FPR, and the empirical $ARL_0$ serve as secondary indicators of overall monitor stability, not specific to the junction.

\begin{table}[t]
\centering
\resizebox{\linewidth}{!}{
\begin{tabular}{lcccc}
\toprule
\multirow{2}{*}{Dataset} & \multicolumn{2}{c}{Heterog.} & \multicolumn{2}{c}{Control} \\
\cmidrule(lr){2-3} \cmidrule(lr){4-5}
 & Pairs & Junct.-FPR & Pairs & Junct.-FPR \\
\midrule
KITTI & 24 & 0.0 & 18 & 0.0 \\
ZOD   & 30 & 0.0 & 30 & 0.0 \\
ONCE  & 40 & 0.0 & 32 & 0.0 \\
\bottomrule
\end{tabular}
}
\caption{Junction-FPR for the density-based scene-change experiment. Heterogeneous pairs concatenate a dense and a sparse sequence; control pairs concatenate two sequences of the same density group.}
\label{tab:supp-scene-density}
\end{table}

\begin{table}[t]
\centering
\resizebox{\linewidth}{!}{
\begin{tabular}{lcccccc}
\toprule
\multirow{2}{*}{Dataset} & \multicolumn{2}{c}{FP/h} & \multicolumn{2}{c}{Global FPR} & \multicolumn{2}{c}{Emp. $ARL_0$ [s]} \\
\cmidrule(lr){2-3} \cmidrule(lr){4-5} \cmidrule(lr){6-7}
 & Het. & Ctrl. & Het. & Ctrl. & Het. & Ctrl. \\
\midrule
KITTI & 3.92 & 3.14 & .250 & .111 & 919 & 1,146 \\
ZOD   & 2.79 & 3.78 & .233 & .400 & 1,292 & 952 \\
ONCE  & 0.49 & 0.56 & .275 & .312 & 7,292 & 6,454 \\
\bottomrule
\end{tabular}
}
\caption{Secondary stability metrics for the density-based scene-change experiment (heterogeneous vs.\ control pairs).}
\label{tab:supp-scene-density-secondary}
\end{table}

\smallskip
\noindent\textbf{Results.} The junction-FPR is $0.0\%$ on all three datasets, for both heterogeneous and control pairs (Table~\ref{tab:supp-scene-density}): the concatenated sequences do raise alarms overall, at FP/h and empirical $ARL_0$ levels consistent with the clean-sequence results of Table~\ref{table:main-results-expanded} (Table~\ref{tab:supp-scene-density-secondary}), but none fall near the density change, indicating ordinary clean-sequence non-stationarity rather than the scene transition itself. We note, however, that pairs are formed from a shared pool of sequences and are thus not fully independent observations.

\begin{table}[t]
\centering
\resizebox{\linewidth}{!}{
\begin{tabular}{lcccc}
\toprule
\multirow{2}{*}{Variant} & \multicolumn{2}{c}{Heterog.} & \multicolumn{2}{c}{Control} \\
\cmidrule(lr){2-3} \cmidrule(lr){4-5}
 & Pairs & Junct.-FPR & Pairs & Junct.-FPR \\
\midrule
ZOD country/daypart & 30 & 0.0 & 30 & 0.0 \\
ONCE weather/period  & 4  & 0.0 & 2  & 0.0 \\
\bottomrule
\end{tabular}
}
\caption{Junction-FPR for the metadata-based scene-change variants. The ONCE variant covers very few pairs and lacks adequate coverage of conditions such as rain or night driving.}
\label{tab:supp-scene-metadata}
\end{table}

The metadata-based variants (Table~\ref{tab:supp-scene-metadata}) corroborate this picture on ZOD, with $0.0\%$ junction-FPR for both heterogeneous and control pairs; the ONCE metadata variant is too small (4 vs.\ 2 pairs) to be informative on its own. Overall, within the scene changes covered by this experiment, MiCaDe's alarms do not concentrate on the scene transition.

\subsection{Robustness of 2D--3D Matching under Miscalibration}
\label{sec:supp-matching-robustness}

As discussed in Section~\ref{sec:late-fusion}, the Bounding Box Matching procedure operates on projected 3D boxes, and is therefore itself affected by the calibration error it is meant to help monitor: a sufficiently large miscalibration could in principle produce wrong 2D--3D correspondences, which would corrupt the geometric consistency metrics and reduce detection power. We evaluate this effect directly using ground-truth object identities. For each match, we say it is \emph{accurate} if the projected 3D detection and the 2D detection are associated with the same annotated object in both modalities; otherwise, the match is counted as an incorrect correspondence.

We compare matching accuracy and the average number of retained matches per frame on ONCE, before and after an abrupt calibration change (Table~\ref{tab:supp-matching-robustness}). Matching accuracy decreases only marginally under miscalibration, showing that the assignment procedure remains largely reliable even after the calibration drift. The main effect of miscalibration is instead a reduction in the number of valid matches retained per frame, as fewer projected boxes reach the IoU threshold $\rho$. Since MiCaDe accumulates evidence over batches rather than relying on any individual match, this reduction primarily lengthens the detection delay rather than the detection power, consistently with the results reported in Section~\ref{sec:experiments}.

\begin{table}[t]
\centering
\resizebox{\linewidth}{!}{
\begin{tabular}{lcc}
\toprule
Metric & Before miscalibration & After miscalibration \\
\midrule
Matching accuracy & 92\% & 89\% \\
Matches per frame & 5 & 3 \\
\bottomrule
\end{tabular}
}
\caption{Effect of an abrupt calibration change on 2D--3D matching, on ONCE. Matching accuracy is only marginally affected, while the average number of retained matches per frame decreases.}
\label{tab:supp-matching-robustness}
\end{table}

\subsection{Robustness to Sensor Degradation}
\label{sec:supp-degradation}

Since MiCaDe relies on object detections extracted from the LiDAR stream, a natural concern is whether a degradation of the LiDAR sensor itself, rather than a calibration drift, could be mistaken for miscalibration. We evaluate this on KITTI Raw, considering two families of LiDAR degradation: \emph{beam loss}, where the number of vertical scan rings is reduced from 64 to 32, 16, or 8, and \emph{sector loss}, where a $45^\circ$ or $90^\circ$ azimuthal sector of the scan is entirely removed, simulating a localized sensor occlusion or fault.

\smallskip
\noindent\textbf{Negative control.} As in Section~\ref{sec:supp-scene-changes}, calibration remains correct throughout while the LiDAR degrades from a known onset; we report FP/h together with the \emph{onset ratio}, the fraction of alarms falling within $\pm 15$ batches of the onset (n/a if no alarm is raised at all).

\begin{table}[t]
\centering
\begin{tabular}{lcc}
\toprule
Degradation & FP/h & Onset ratio \\
\midrule
64$\to$32 rings & 1.28 & 0.00 \\
64$\to$16 rings & 3.87 & 0.00 \\
64$\to$8 rings  & 5.52 & 0.00 \\
$45^\circ$ sector missing & 0.00 & n/a \\
$90^\circ$ sector missing & 0.00 & n/a \\
\bottomrule
\end{tabular}
\caption{Negative control for LiDAR degradation on KITTI Raw at $ARL_0=10000$: calibration remains correct throughout, only the LiDAR stream is degraded from a known onset.}
\label{tab:supp-degradation-control}
\end{table}

At $ARL_0=10000$, no false alarm falls within the window for any degradation level -- including a fully missing $90^\circ$ sector, arguably the most severe case tested -- confirming that degradation is not confused with calibration drift (with only $0.774$~h of aggregated exposure, the empirical $ARL_0$ here is censored, as in Section~\ref{sec:experiments}).

\smallskip
\noindent\textbf{Sensitivity under concurrent miscalibration.} We then inject a genuine calibration drift on top of the degraded stream, using a three-phase protocol: clean segment, degradation onset at $15\%$, drift onset at $35\%$.

\begin{table}[t]
\centering
\begin{tabular}{lccc}
\toprule
Degradation & FNR & FPR & Delay \\
\midrule
64$\to$32 rings & .207 & .000 & 12.3~s \\
64$\to$16 rings & .379 & .000 & 13.9~s \\
64$\to$8 rings  & .600 & .040 & 18.2~s \\
$45^\circ$ sector missing & .207 & .000 & 12.7~s \\
$90^\circ$ sector missing & .207 & .000 & 13.4~s \\
\bottomrule
\end{tabular}
\caption{Detection of a calibration drift injected on top of an already-degraded LiDAR stream, on KITTI Raw, at $ARL_0=10000$.}
\label{tab:supp-degradation-sensitivity}
\end{table}

At $ARL_0=10000$, the FPR stays (almost) zero across all degradation levels. Sensitivity remains good under moderate beam loss and either sector loss, but degrades markedly at only 8 of 64 rings (FNR $0.600$, FPR no longer zero); this rise in FNR is driven by the LiDAR object detector itself, which becomes less robust and produces fewer valid matches under such severe degradation, rather than by a failure of the monitoring scheme.

\smallskip
\noindent\textbf{Take-away.} Sensor degradation and calibration drift are largely distinguishable by MiCaDe: even a full $90^\circ$ sector loss is never confused with miscalibration, and moderate beam loss does not compromise detection of a concurrent drift. Performance only degrades under severe beam loss, an effect bounded by the object detector's own robustness to sparse input rather than by the monitoring scheme itself.

\section{Real-World Dataset}
\label{sec:dataset-supplementary}

\begin{figure}[t]
\centering
\includegraphics[width=\linewidth]{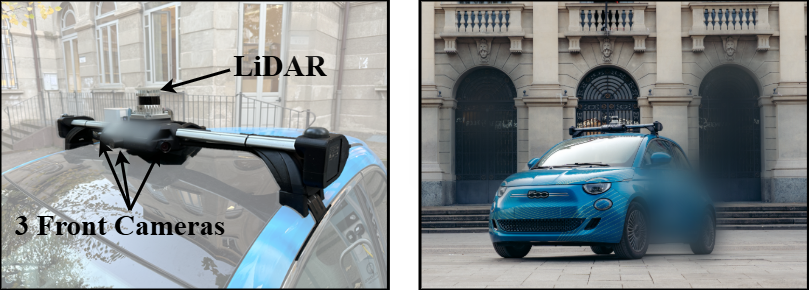}
\caption{The experimental vehicle used for data collection.}
\label{fig:aida-setup}
\end{figure}

\begin{figure}[t]
    \centering
      \begin{subfigure}[t]{.39\linewidth}
        \centering\includegraphics[width=\linewidth]{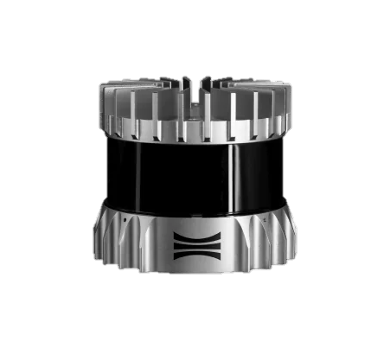}
        \caption{LiDAR: Ouster OS1.}
      \end{subfigure}
      \begin{subfigure}[t]{.6\linewidth}
        \centering\includegraphics[width=\linewidth]{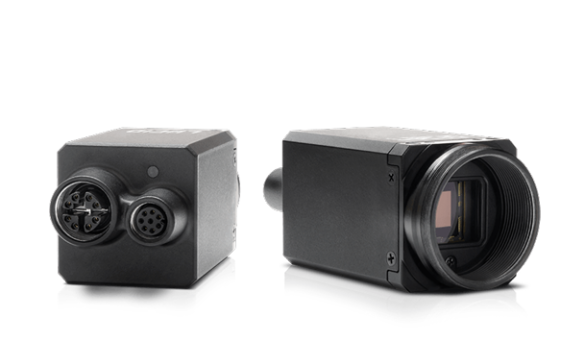}
        \caption{RGB Camera: Lucid Triton 5.4MP.}
      \end{subfigure}
    \caption{Sensors mounted on top of the vehicle used to collect the data.}
    \label{fig:aida-sensors}
\end{figure}

\begin{figure}[t]
\centering
\includegraphics[width=\linewidth]{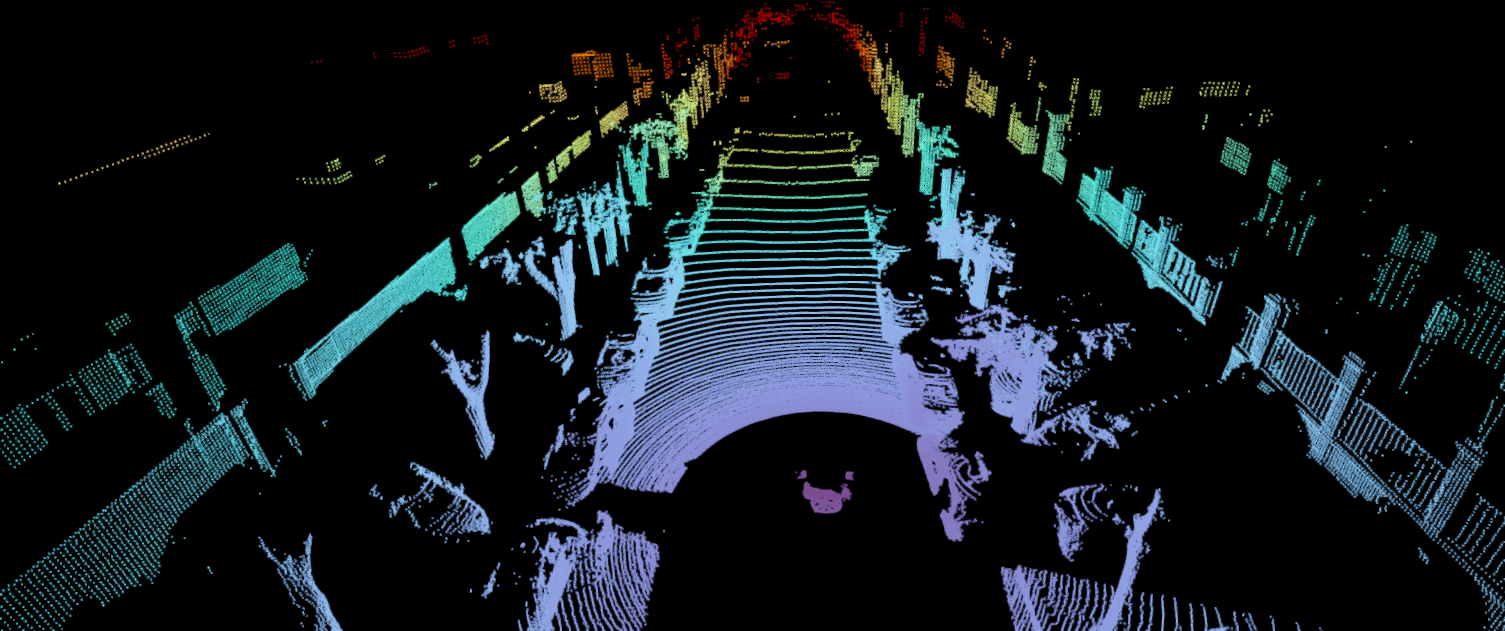}
\caption{An example of point cloud taken from the Ouster OS1 LiDAR.}
\label{fig:aida-point-cloud}
\end{figure}

\begin{figure}[t]
\centering
\includegraphics[width=\linewidth]{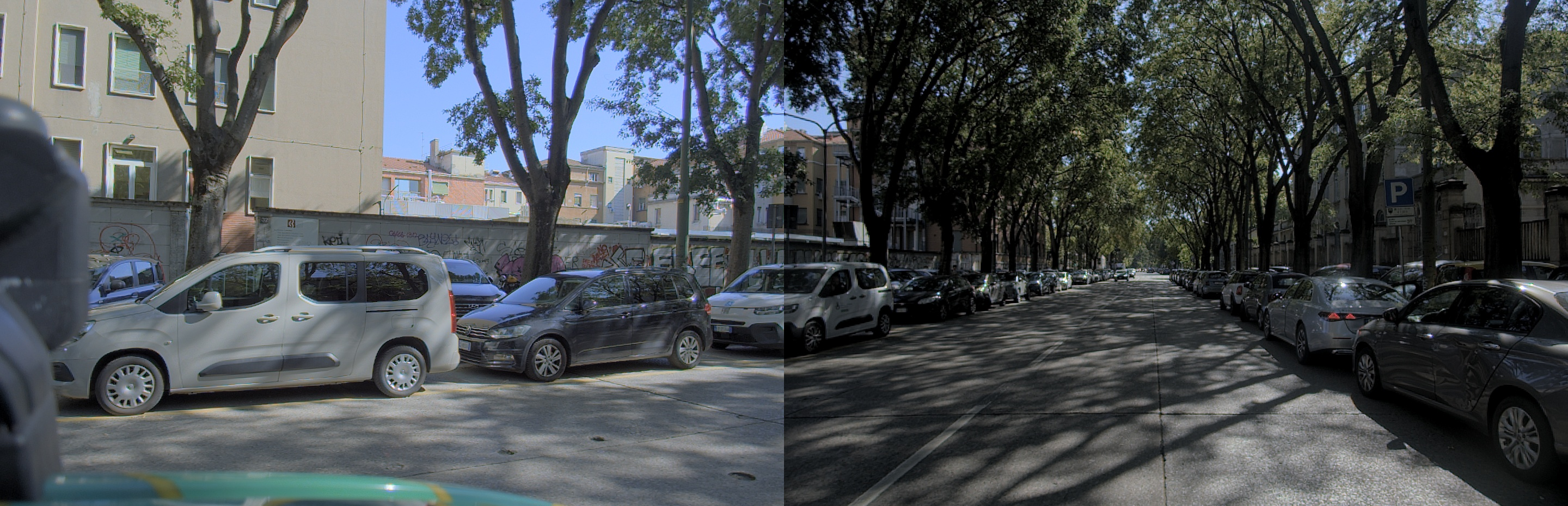}
\caption{An example of front and left images taken from the RGB cameras.}
\label{fig:aida-image}
\end{figure}

Our real-world evaluation is conducted on a custom dataset collected using the sensors shown in Figure~\ref{fig:aida-sensors}. In this section, we provide further details with respect to the ones presented in Section \ref{sec:real-world}.
The sensing suite for data collection comprises:
\begin{itemize}
    \item A \textbf{128-beam Ouster OS1} LiDAR, operating at 10~Hz, with a
    $360^\circ$ horizontal field of view, $45^\circ$ vertical field of view. Figure~\ref{fig:aida-point-cloud} shows an example of an OS1-128 LiDAR scan.

    \item Three \textbf{Lucid Triton 5.4\,MP} RGB cameras,
    mounted as front-left, front-center, and front-right; each operating at
    30~Hz with a resolution of $1440{\times}720$ pixels. Figure~\ref{fig:aida-image} shows an example of front-left and front-center RGB images.
\end{itemize}

All sensors are hardware-timestamped but run at different frame rates. To build consistent multimodal frames, we perform temporal alignment by matching each LiDAR scan with the closest RGB images from all cameras. Pairs whose timestamps differ by more than 0.02\,s are discarded. This procedure produces a stream of temporally synchronized multimodal frames.

\begin{figure}[t]
\centering
\includegraphics[width=\linewidth]{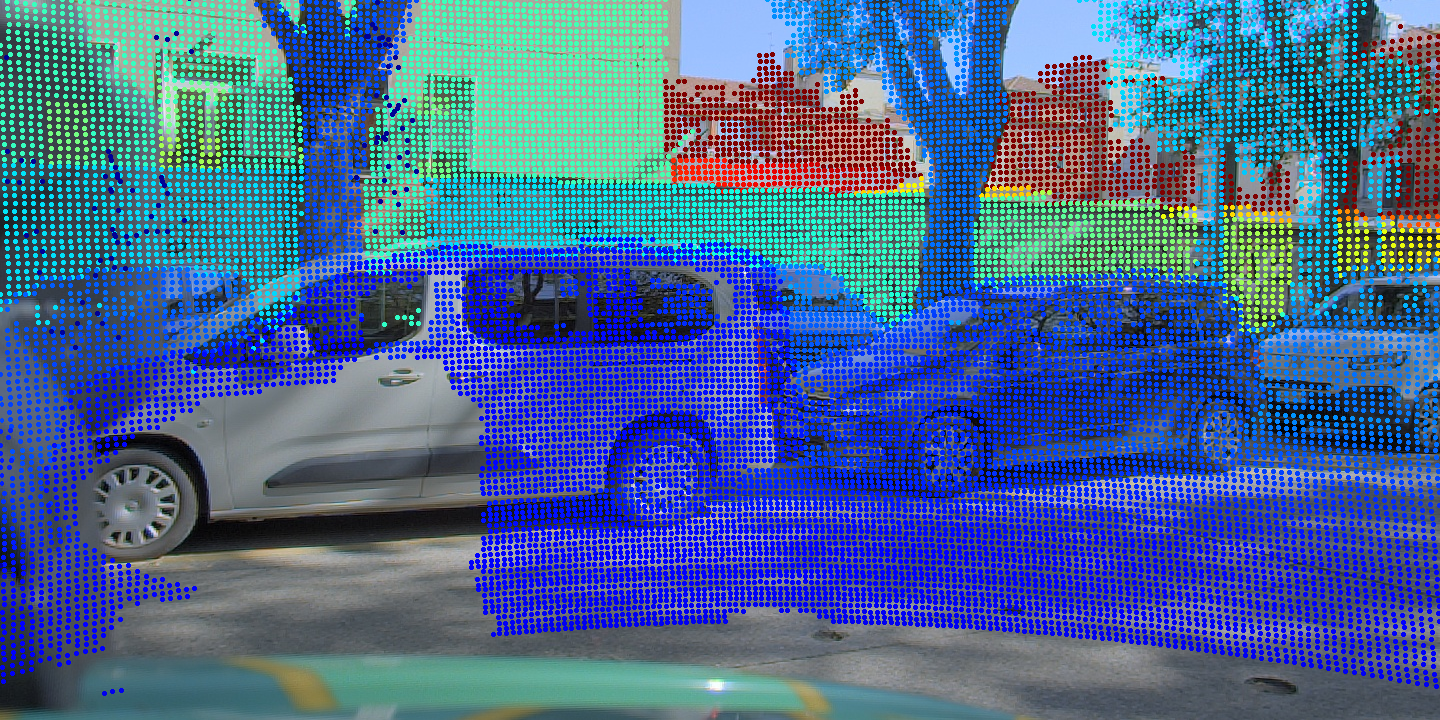}
\caption{An example of a calibrated frame. The projection of the points is precisely inside the objects, up to small errors due to temporal synchronization.}
\label{fig:aida-calibrated}
\end{figure}

\begin{figure}[t]
\centering
\includegraphics[width=\linewidth]{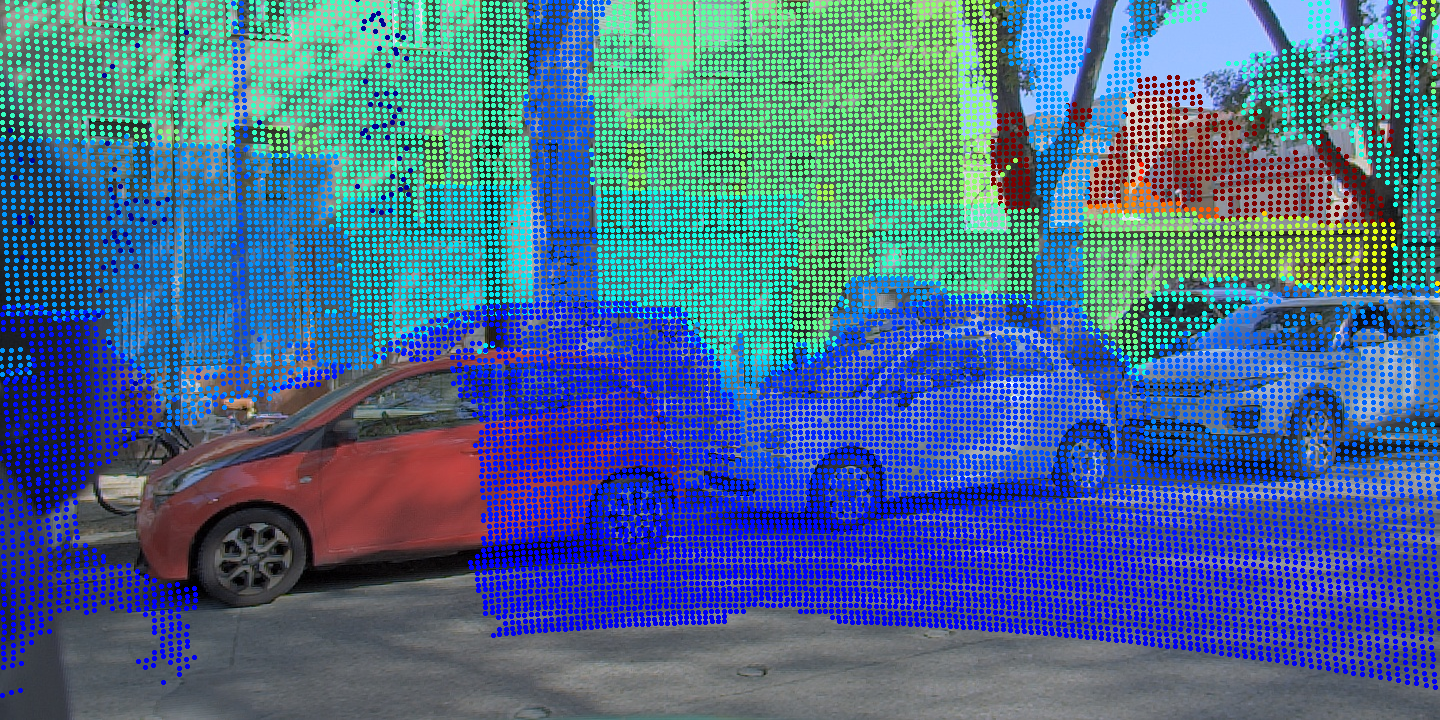}
\caption{An example of an uncalibrated frame. The projection of the points is clearly inconsistent, which is visible from the red car in front of the camera.}
\label{fig:aida-uncalibrated}
\end{figure}

\paragraph{Data Collection Protocol.}
The dataset contains approximately one hour of real urban driving, collected in residential and light-traffic areas. The vehicle follows a standard route including: straight roads, intersections, slow traffic segments, and mild accelerations/braking. This ensures diverse distributions of object distances, orientations, and occlusion patterns.

During the first half of the recording, all sensors are precisely calibrated, and the system is operated in nominal conditions. During the second half, we intentionally introduce a controlled misalignment on the front-left camera:
\begin{itemize}
    \item rotation of $3^\circ$ in roll, $3^\circ$ in pitch, and $3^\circ$ in yaw;
    \item no induced translation shift (only rotational perturbation), matching common real-world mechanical drift.
\end{itemize}
The perturbation is physically introduced by loosening and rotating the camera mount before tightening it again. As a result, only rotational miscalibrations can be induced in a controlled and repeatable manner, whereas introducing a pure translational offset would require physically relocating the sensor and redesigning the mounting system. Once rotated, the camera remains rigidly fixed, so the induced drift is constant throughout the miscalibrated portion of the sequence.

This procedure yields a total of 2151 calibrated frames and 5234 miscalibrated frames. An example calibrated frame is shown in Figure~\ref{fig:aida-calibrated}, while Figure~\ref{fig:aida-uncalibrated} shows a frame after miscalibration, where projected LiDAR points appear systematically shifted with respect to object edges.

\paragraph{Reference Frames for the GMM.}
To build the reference distribution used by the change detector, we select the first 1000 calibrated frames, using only those with at least one valid 3D-2D detection match. These frames define the nominal geometric consistency statistics for the sensor configuration.

\paragraph{Usage in the Experiments.}
After the initial 1000 reference frames, the full MiCaDe pipeline is applied to the remaining calibrated portion and the entire miscalibrated portion. We preserve natural traffic conditions and do not inject synthetic noise,
ensuring that the evaluation reflects real detection variability, occlusions,
and motion blur. Figure~\ref{fig:metrics-aida-full} reports the evolution of the complete set of monitored metrics on this sequence, extending the three metrics shown in Figure~\ref{fig:metrics-aida} of the main paper.

\begin{figure*}[t]
\centering
\includegraphics[width=\linewidth]{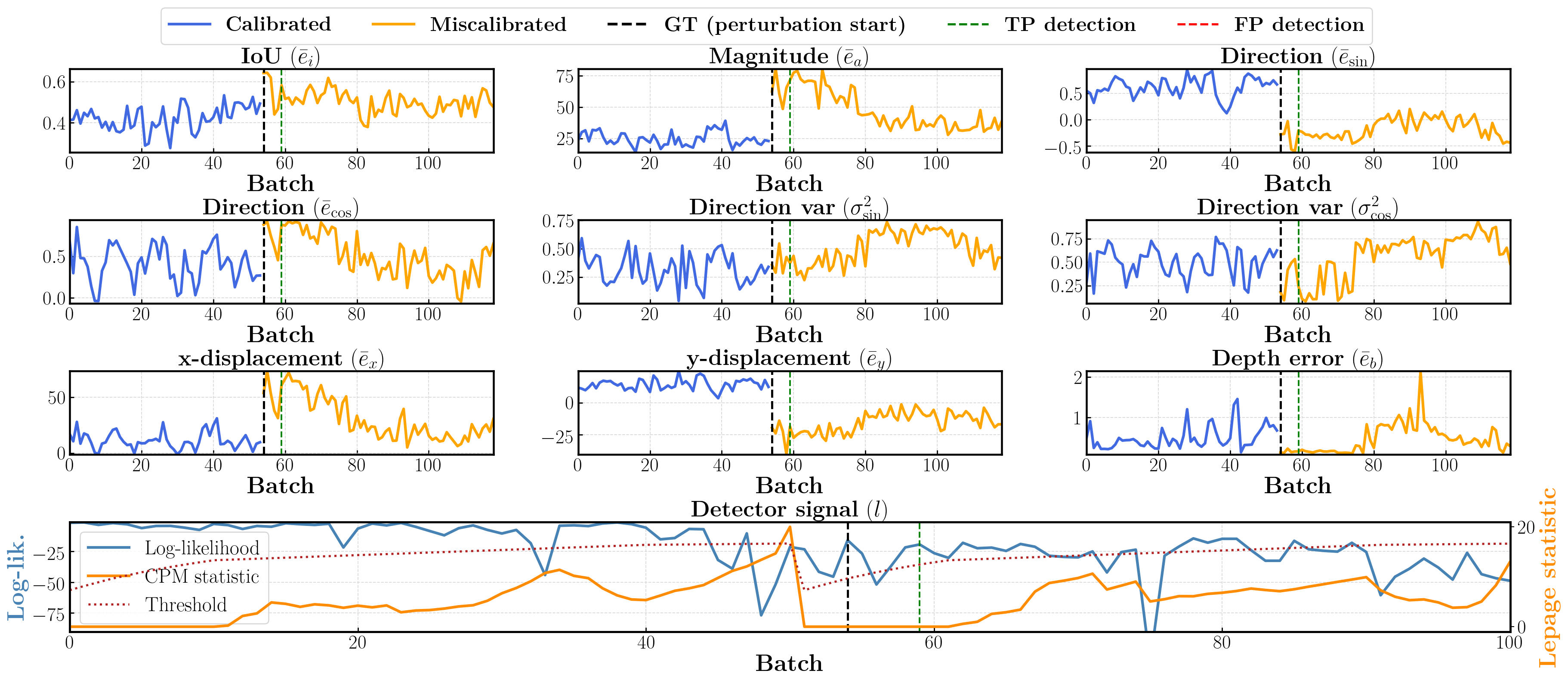}
\caption{
Evolution of the complete set of monitored metrics and detector signal on the real-world sequence.
The black dashed line marks the manually induced calibration change, while the green line indicates the detection raised by MiCaDe.
Following the perturbation, all metrics exhibit visible shifts, including the center displacement terms ($e_x$, $e_y$), the displacement magnitude ($e_a$), and the directional indicators ($e_{\sin}$, $e_{\cos}$, $\sigma^2_{\sin}$, $\sigma^2_{\cos}$).
The bottom panel reports the log-likelihood stream and the corresponding CPM statistic, which crosses the decision threshold shortly after the perturbation, correctly identifying the calibration drift.
}
\label{fig:metrics-aida-full}
\end{figure*}

\paragraph{Release.} The dataset, including the raw multimodal recordings and the known calibration ground truth, will be made publicly available upon acceptance of the paper.

\section{Qualitative Change-Detection Examples}
\label{sec:supp-qualitative}

Figures~\ref{fig:supp-kitti-tp}--\ref{fig:supp-once-fn} provide qualitative examples of MiCaDe on KITTI Raw, ZOD, and ONCE. Each example reports the temporal evolution of the monitored metrics before and after the injected calibration change, together with two representative image frames. Blue curves correspond to calibrated observations, orange curves to observations after the perturbation, the black dashed line marks the ground-truth change point, and the green dashed line indicates the alarm raised by MiCaDe. The bottom images show the geometric relationship between 2D detections (green boxes), projected 3D boxes (red dashed boxes), and the displacement vectors used to compute the monitored consistency metrics.

The true-positive examples in Figures~\ref{fig:supp-kitti-tp}, \ref{fig:supp-zod-tp}, and~\ref{fig:supp-once-tp} show that calibration drift typically induces coherent changes across several metrics. In particular, the IoU decreases after the perturbation, while the displacement magnitude and center-offset components increase. Directional metrics also become more structured, reflecting the emergence of a preferred displacement direction between projected LiDAR boxes and image detections. These changes are visually consistent with the image examples, where the projected 3D boxes become systematically shifted with respect to the corresponding 2D detections after the perturbation.

The examples also illustrate the difference between delay measured in batches and delay measured in frames. MiCaDe operates on batches of $n=15$ matched detections, not on individual frames. Therefore, the conversion from batch delay to frame delay depends on how many valid 3D--2D correspondences are available in the scene. In dense scenes with many detected objects, batches are filled quickly and a small batch delay corresponds to only a few frames. This is visible in the ZOD example, where the change is detected in the same batch as the perturbation and corresponds to a delay of only 3 frames. Similarly, the KITTI example is detected after 1 batch, corresponding to 4 frames, while the ONCE true-positive example is detected after 4 batches, corresponding to 14 frames.

Figure~\ref{fig:supp-once-fn} shows a representative false-negative case on ONCE. In this sequence, the injected perturbation is very small: the rotational offsets are close to zero and the translations are below $0.1$~m. As a consequence, the post-change metrics remain close to the variability observed during the calibrated portion of the sequence. The IoU, displacement magnitude, directional components, and depth error do not exhibit a persistent shift strong enough to separate the miscalibrated regime from nominal fluctuations. Visually, the projected 3D boxes remain only mildly misaligned with the 2D detections. This example highlights a natural limitation of object-level calibration monitoring: when the induced calibration drift is too small to produce a measurable change in geometric consistency, the sequential detector may not accumulate sufficient evidence to raise an alarm.

Overall, these qualitative examples confirm the behaviour observed quantitatively in the main paper. MiCaDe detects calibration changes quickly when the perturbation produces a consistent geometric signature across matched detections, while failures mainly occur for very small perturbations whose effect is comparable to natural detection noise, scene variability, and matching uncertainty.

\begin{figure*}[t]
\centering
\includegraphics[width=0.9\linewidth]{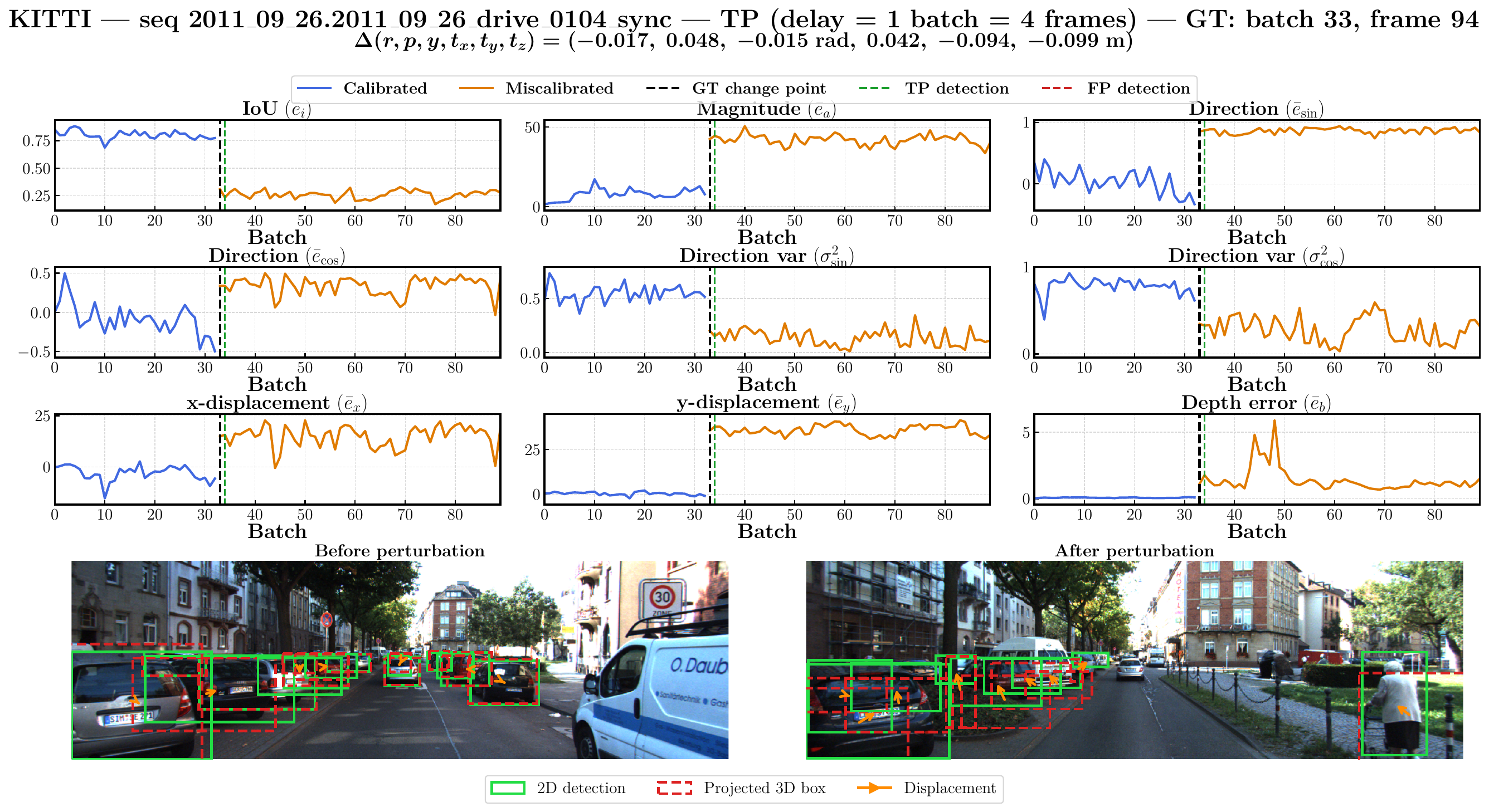}
\caption{Qualitative true-positive example on KITTI Raw. The perturbation is detected after 1 batch, corresponding to 4 frames.}
\label{fig:supp-kitti-tp}
\end{figure*}

\begin{figure*}[t]
\centering
\includegraphics[width=0.9\linewidth]{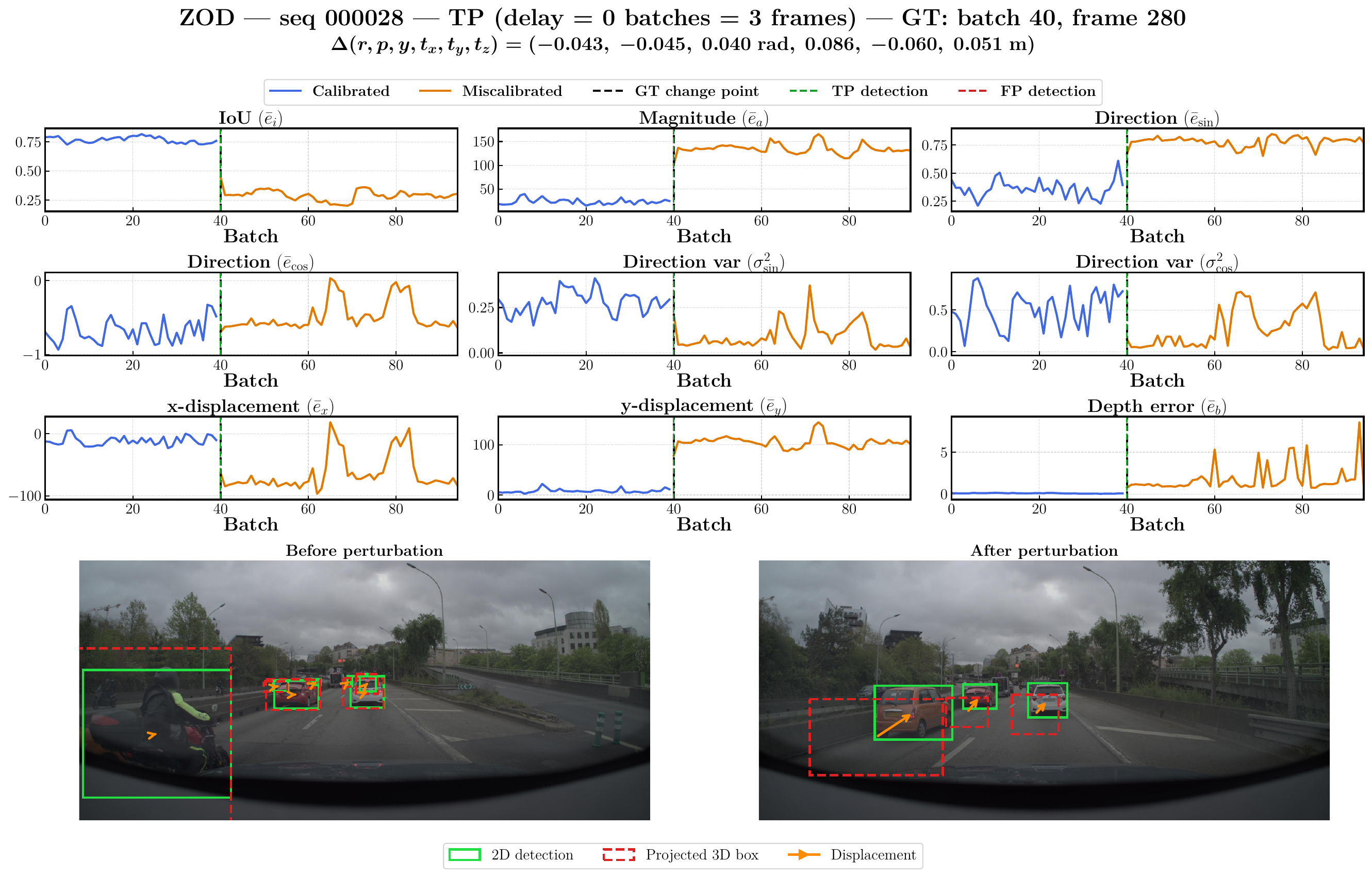}
\caption{Qualitative true-positive example on ZOD. MiCaDe detects the perturbation in the same batch, corresponding to a delay of 3 frames.}
\label{fig:supp-zod-tp}
\end{figure*}

\begin{figure*}[t]
\centering
\includegraphics[width=0.9\linewidth]{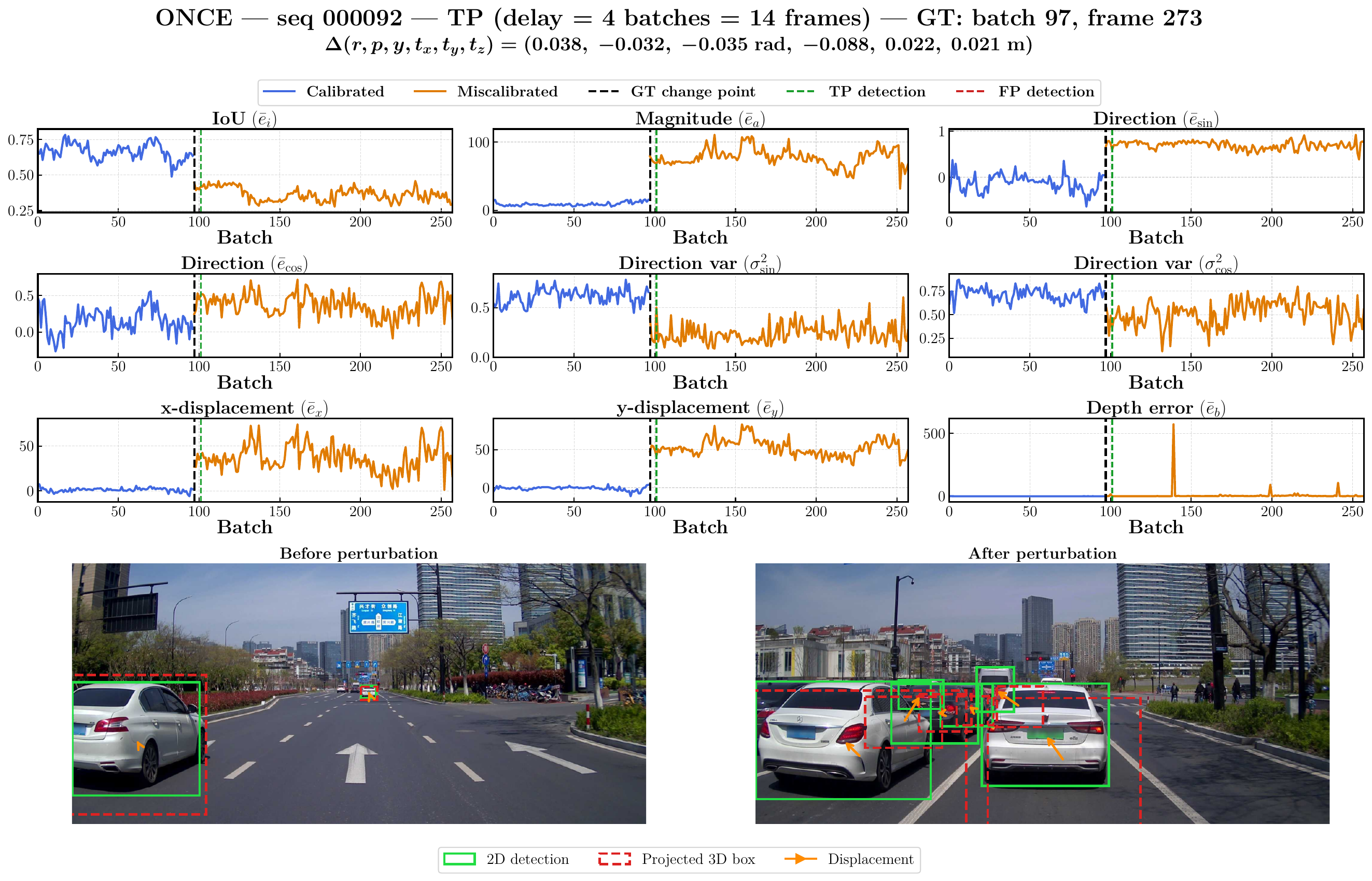}
\caption{Qualitative true-positive example on ONCE. The perturbation is detected after 4 batches, corresponding to 14 frames.}
\label{fig:supp-once-tp}
\end{figure*}

\begin{figure*}[t]
\centering
\includegraphics[width=0.9\linewidth]{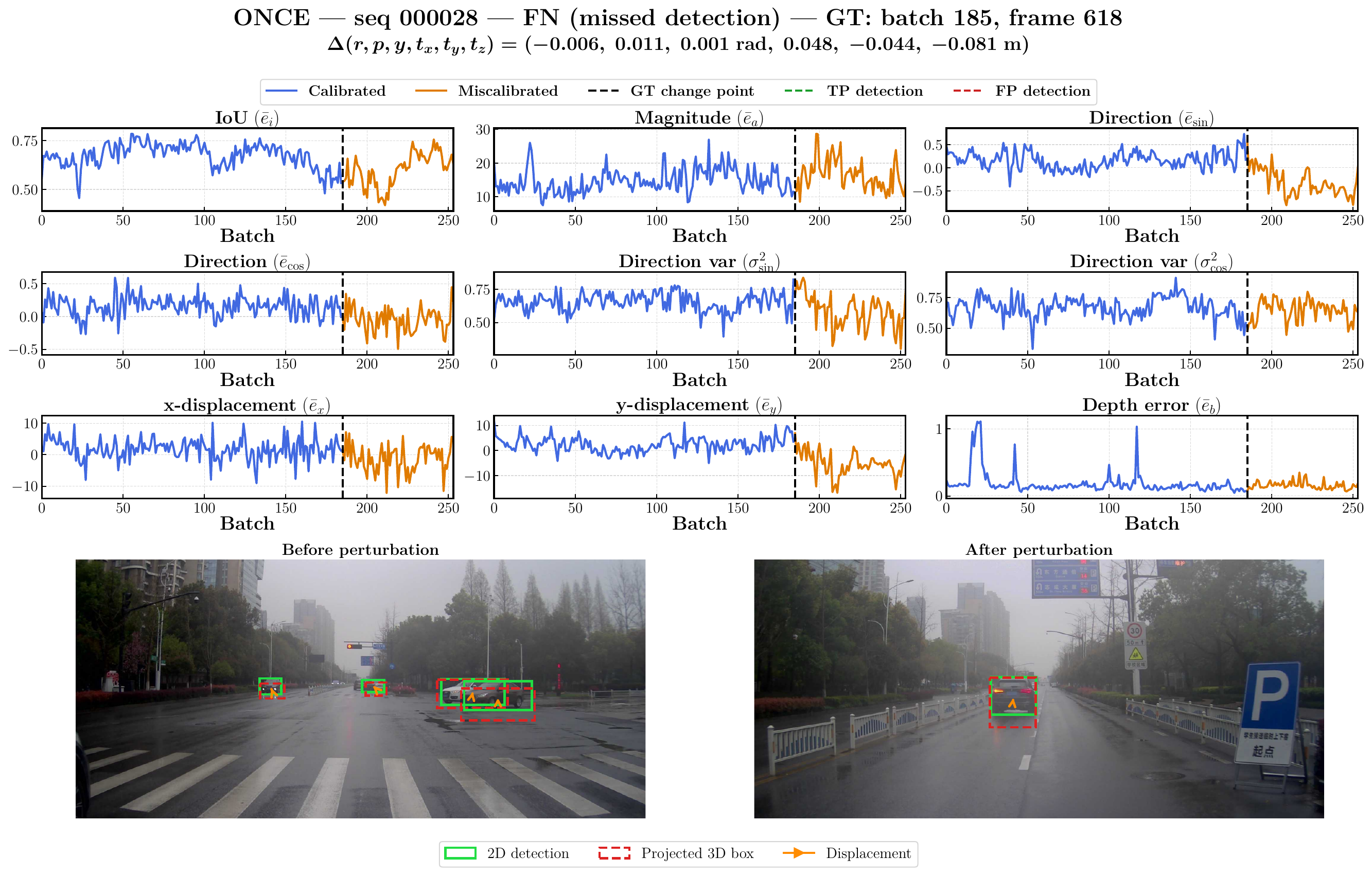}
\caption{Qualitative false-negative example on ONCE. The perturbation is very small and produces only mild changes in the monitored metrics, preventing MiCaDe from raising an alarm.}
\label{fig:supp-once-fn}
\end{figure*}

\end{document}